\documentclass[letterpaper]{article} 
\usepackage{aaai2027}  
\usepackage[hyphens]{url}  
\usepackage{graphicx} 
\usepackage{natbib}  
\usepackage{caption} 
\usepackage{graphicx}
\usepackage{booktabs}

\usepackage{tcolorbox}
\tcbuselibrary{breakable}
\usepackage{amsmath,amsfonts}
\usepackage{algorithm}
\usepackage{algpseudocode}
\usepackage{array}
\usepackage{textcomp}
\usepackage{pifont}
\usepackage{url}
\usepackage{verbatim}
\usepackage{graphicx}
\usepackage{fancybox}
\usepackage[table]{xcolor}
\usepackage{xcolor}
\usepackage{arydshln}  
\usepackage{booktabs}       
\usepackage{amsfonts}       
\usepackage{nicefrac}       
\usepackage{microtype}      
\usepackage[dvipsnames]{xcolor}         
\usepackage{enumitem}
\usepackage{multirow}
\usepackage{amsmath}
\usepackage{mdframed}
\usepackage{subcaption}
\usepackage{soul}
\usepackage{amssymb}
\usepackage{pifont}

\usepackage{newfloat}
\usepackage{listings}
\DeclareCaptionStyle{ruled}{labelfont=normalfont,labelsep=colon,strut=off} 
\floatstyle{ruled}
\newfloat{listing}{tb}{lst}{}
\floatname{listing}{Listing}

\usepackage{booktabs}

\title{ExpertVerse: A General-Purpose Benchmark for Expert-Level Reasoning in Knowledge-Intensive Visual Synthesis
}
\author
{
Yuan Wang\textsuperscript{\rm 1,2}, Yongchao Du\textsuperscript{\rm 1}, Mengting Chen\textsuperscript{\rm 1$\dagger$}, Jinsong Lan\textsuperscript{\rm 1}, Jiaxuan Luo\textsuperscript{\rm 3} \\ Xuetao Feng\textsuperscript{\rm 2}, Xiaoyong Zhu\textsuperscript{\rm 1}
}

\affiliations{
\textsuperscript{\rm 1}Alibaba Group \quad
\textsuperscript{\rm 2}Tsinghua University \quad
\textsuperscript{\rm 3}Johns Hopkins University‌ \quad
 $\dagger$Corresponding Author.
}

\begin{document}

\maketitle

\begin{abstract}
Recent advances in multimodal generative models have enabled instruction-based image generation to move beyond semantic manipulation to knowledge-driven visual reasoning. However, these methods focus on explicit commonsense reasoning, shallow causal understanding, and direct knowledge recall, failing at knowledge-intensive generation. We develop \textbf{ExpertVerse}, a capability-centric benchmark to evaluate generative models via knowledge-intensive lens. ExpertVerse stratifies reasoning generation across an orthogonal taxonomy of \textit{9 cognitive capabilities} and \textit{8 expert disciplines}, yielding \textit{58 sub-disciplines}. We curate 1,611 expert-annotated instances covering single-image editing, multi-image composition, and text-to-image generation. We further develop an automated workflow to produce \textbf{ExpertVerse-100K}, a large-scale dataset with reasoning traces and knowledge-anchored rationale annotations. Based on this, we train \textbf{KnowThinker} with RL fine-tuning, a VLM reasoning engine with world knowledge that jointly generates thinking processes and refined instructions. Towards the cross-modal credit misalignment and multi-objective gradient conflicts in multi-reward optimization, we propose a tailored Bootstrapped Pareto Policy Optimization (BPPO), which synergizes Bootstrapping Reward Rectification (BRR) and Conflict-Aware Pareto Advantage Fusion (CPAF). Extensive results of both open-source and proprietary models exposes critical reasoning deficits, highlighting imperative for knowledge-intensive benchmarks towards next-generation visual generation.
\end{abstract}

\section{Introduction}
Recently, instruction-based image synthesis~\cite{zhao2024ultraedit,brooks2023instructpix2pix,chen2024zero} has evolved from basic semantic manipulation to complex visual reasoning, driven by the emergent capabilities of vision-language models (VLMs)~\cite{bai2025qwen3,qwen35blog} and generative models~\cite{deng2025emerging,batifol2025flux,comanici2025gemini,liu2026continuous}, driving advancements in virtual try-on~\cite{chen2026tstars,sun2026tryoncrafter} and avatar creation~\cite{xi2025depthdance,wang2024holistic,song2026fashionchameleon,liu2026improving,chen2024zero,chen2024livephoto}. Early image editing benchmarks~\cite{liu2025step1x,ye2026imgedit} primarily target attribute-level modifications. Beneath compelling visual fidelity lies an inability to enforce physical plausibility and logical coherence when implicit reasoning is commonly required. To bridge this gap, recent reasoning-aware benchmarks have expanded cognitive evaluation across three dimensions: \textit{reasoning types, knowledge hierarchies, and scenario complexity}. RISEBench~\cite{zhao2026envisioning} pioneers hypothetical reasoning across temporal, spatial, causal, and logical dimensions. KRISBench~\cite{wu2026kris} proposes knowledge-grounded taxonomy assessing factual, conceptual, and procedural knowledge types. UniREditBench~\cite{han2025unireditbench} extends this scope with the game-world scenarios and multi-object interaction tasks. 

Despite effectiveness, they exhibit a bottleneck in benchmarking knowledge-aware reasoning generation: (1) they largely center on explicit commonsense, shallow causality, and direct knowledge recall (\textit{e.g.}, \textit{melting ice} or \textit{burning candles}), leaving model generalization to\textbf{ knowledge-intensive reasoning scenarios} (\textit{e.g.}, \textit{nutritional analysis} and \textit{historical-figure social media}) insufficiently examined. (2) KRIS-Bench primarily adopts a task-centric knowledge typology, lacking a \textbf{universal, capability-centric taxonomy} to assess the cross-disciplinary reasoning capabilities of image editing models (see Tab.~\ref{tab:compared with other bmk}). (3) They are largely restricted to single-image manipulation, with limited coverage of \textbf{multi-image composition} and \textbf{text-to-image synthesis} for comprehensive generation evaluation. A holistic benchmark to evaluate knowledge-intensive generation abilities across diverse knowledge dimensions has thus become increasingly essential.

\begin{figure*}[h]
    \centering
    \vspace{-0.4cm}
    \captionsetup{skip=2pt}
    \caption{
    ExpertVerse includes 8 expert disciplines covering single-image editing, multi-image composition, and T2I generation.}
    \includegraphics[width=1\textwidth]{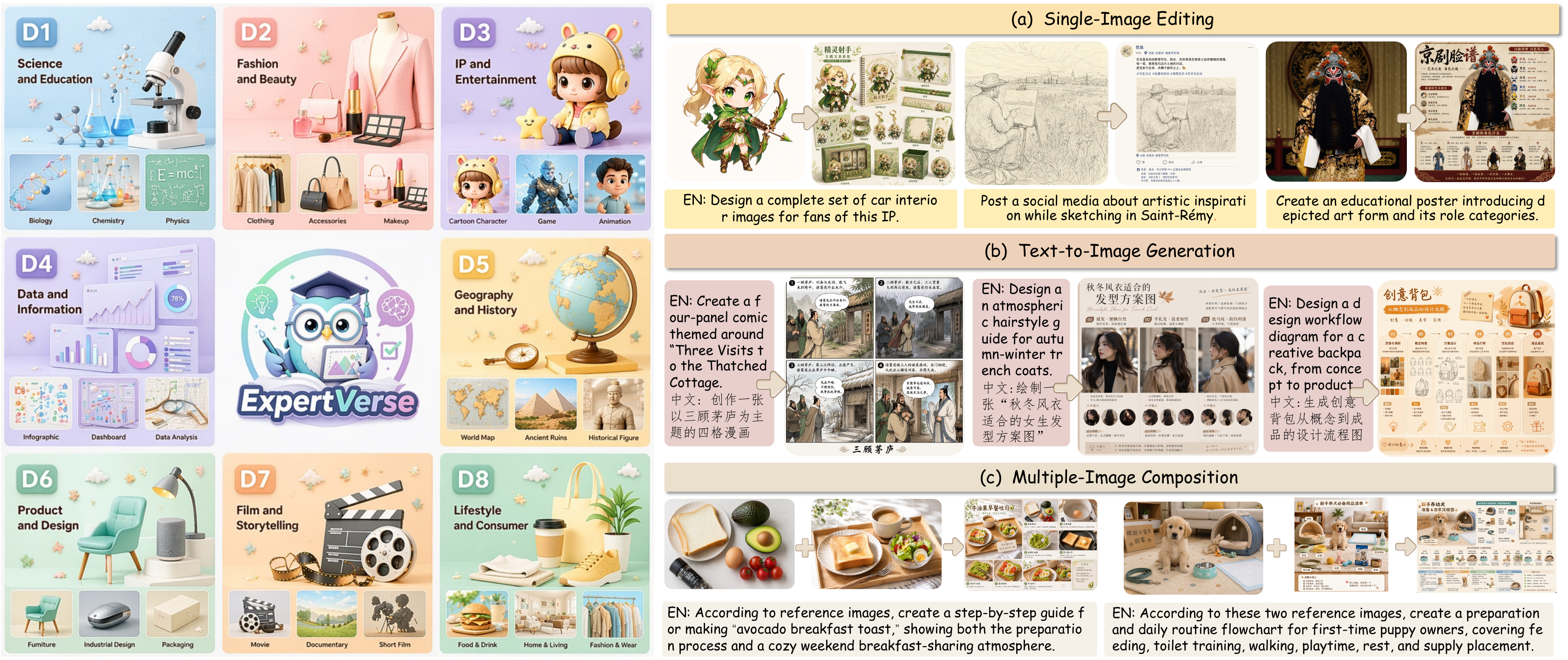}
    \label{fig:teaser}
    \vspace{-0.24cm}
\end{figure*}

To this end, we develop ExpertVerse, a capability-centric benchmark to evaluate visual synthesis models via knowledge-intensive lens. In Fig.~\ref{fig:teaser} and Tab.~\ref{tab:compared with other bmk}, ExpertVerse stratifies tasks along an orthogonal taxonomy of \textit{9 cognitive capabilities} and \textit{8 expert categories} (\textit{e.g.}, science, history, information, life, etc.), yielding \textit{58 sub-categories} that ensure broad generality and high knowledge density. We curate 1,611 expert-annotated instances covering single-image editing, multi-image composition, and text-to-image synthesis. We initiate data synthesis with handcrafted seed reference text prompts—comprising image descriptions, user instructions, and domain-specific rationales—which are subsequently scaled via a proprietary VLM. To maximize the diversity of ExpertVerse, we devise a hierarchical expansion strategy that further partitions each of \textit{58 orthogonal sub-categories} into multiple atomic generation tasks. Built upon this, we curate ExpertVerse-100K, a large-scale dataset for knowledge-intensive image generation with step-wise CoT traces and knowledge-anchored rationale annotations generated using multimodal foundation model. 

\begin{table*}[]
\centering
\small
\resizebox{\textwidth}{!}{%
\begin{tabular}{lcccccccccccccc}
\toprule
\multirow{2}{*}{Benchmark} & \multirow{2}{*}{\#Task} & \multirow{2}{*}{size} & \multicolumn{3}{c}{Generation Types} & Knowledge   & Creative & Logical   & Aesthetic & Spatial        & Temporal  & Narrative    & Cross-domain    & Long \\ 
                           &           &            & S-I2I     & M-I2I     & T2I    & Application & design   & reasoning & judgment  & transformation & reasoning & construction & transfer & Text \\ \midrule
SmartEdit       &  7   &   219                    &   \ding{51}        &   \ding{55}        &   \ding{55}     &     \ding{55}        &    \ding{55}      &     \ding{55}      &     \ding{55}      &      \ding{51}          &      \ding{55}     &    \ding{55}          &  \ding{55}        &   \ding{55}    \\
RISEBench       &   16        &   360                    &     \ding{51}      &  \ding{55}         &   \ding{55}     &   \ding{55}          &     \ding{55}     &   \ding{51}        &   \ding{55}        &       \ding{51}         &    \ding{51}       &    \ding{55}          &  \ding{55}        &   \ding{55}   \\
KRISBench          &   22     &  1267                     &   \ding{51}        & \ding{51}          &   \ding{55}     &    \ding{51}         &   \ding{55}       &     \ding{51}      &   \ding{55}        &    \ding{51}            &   \ding{51}        &   \ding{55}           &   \ding{55}       &   \ding{55}   \\
UniREditBench       &   18        &  2700                     &   \ding{51}        &      \ding{55}     &   \ding{55}     &    \ding{51}         &   \ding{55}       &   \ding{51}        &  \ding{55}         &  \ding{51}                &  \ding{51}         &    \ding{55}          &   \ding{55}      &   \ding{55}   \\
WiseEdit                &   31             &   1220    &   \ding{51}        &   \ding{51}     & \ding{55} &   \ding{51}        &    \ding{51}      &  \ding{51}         &    \ding{55}       &    \ding{51}            &    \ding{51}       &   \ding{55}        &   \ding{51}       &   \ding{55}   \\ 
ExpertVerse        &    58       &   1611                    &   \ding{51}        & \ding{51}          &   \ding{51}     &    \ding{51}         &    \ding{51}      &  \ding{51}         &   \ding{51}        &   \ding{51}             &  \ding{51}         &   \ding{51}           &    \ding{51}      &   \ding{51}   \\ \bottomrule

\end{tabular}
}
\vspace{-0.17cm}
\caption{Comparison of open-source knowledge-informed reasoning-based image generation benchmarks.}
\label{tab:compared with other bmk}
\end{table*}

Recent studies~\cite{yang2026dda,li2025editthinker} show isolating VLM reasoner by keeping visual editor frozen effectively targets reasoning bottleneck in think-then-execute pipeline. Under thinker-centric paradigm, we utilize knowledge-aware supervision from ExpertVerse-100K to train KnowThinker, a VLM reasoner that jointly produces explicit thinking processes and refined instructions. Current methods leverage RL to incentivize the knowledge reasoning capabilities of VLM reasoners and incorporate multi-objective rewards (\textit{e.g.}, image reward and CoT reward) for editing policy optimization. While intuitively appealing, it induces cross-modal credit misalignment and multi-objective gradient conflicts. To this dilemma, we propose Bootstrapped Pareto Policy Optimization (BPPO), which synergizes Bootstrapping Reward Rectification (BRR) and Conflict-Aware Pareto Advantage Fusion (CPAF). BRR addresses cross-modal credit misalignment by bootstrapping modality-specific Group Relative Policy Optimization (GRPO)~\cite{guo2025deepseek} baselines with cross-domain rewards, enabling each advantage to perceive paired modality's quality. CPAF eliminates multi-objective gradient conflicts by \textbf{advantage-level pareto aggregation}, avoiding token-level update cancellation caused by conflicting rewards.

Our findings underscores the imperative for this knowledge-intensive benchmark. KnowThinker sets new state-of-the-art records, surpassing existing editing models by a clear margin.

\begin{itemize}
    \item We develop \textbf{ExpertVerse}, the first knowledge-intensive image generation benchmark, stratified into 58 sub-tasks across 9 reasoning abilities and 8 expert disciplines, supported by 1,611 expertly annotated generation instances.
    \item We propose \textbf{KnowThinker} under thinker-centric training paradigm, an expert-level knowledge reasoner that decompose high-level instructions into step-wise generation.
    \item We advance a RL algorithm BPPO for reasoning synthesis, which includes BRR and CPAF towards cross-modal credit misalignment and multi-objective gradient conflicts issues.
\end{itemize}

\section{Related Works}

\noindent \textbf{Reasoning-based Benchmarks for T2I and Image Editing.} Recent T2I benchmarks increasingly evaluate the reasoning capabilities of generative models. WISE~\cite{niu2025wise} probes world knowledge spanning cultural and physical domains, while UniGenBench++~\cite{wang2025unigenbench++} establishes a unified semantic evaluation taxonomy including 10 primary dimensions, \textit{e.g.}, logical reasoning and relational understanding. In image editing protocol, reasoning-based benchmarks like RISEBench~\cite{zhao2026envisioning} assess the temporal, causal, spatial, and logical abilities of editing models. KRISBench~\cite{wu2026kris} introduces a knowledge-grounded taxonomy covering factual, conceptual, and procedural types, yet remains limited in knowledge depth. UniREditBench~\cite{han2025unireditbench} expands scope with game-world scenarios and multi-object interactions. WiseEdit~\cite{pan2026wiseedit} evaluates cognition- and creativity-informed editing via a three-stage cognitive pipeline of awareness, interpretation, and imagination. To bridge this gap, we introduce ExpertVerse targeting expert-level reasoning edit task that demands knowledge-intensive, multi-hop reasoning across professional domains. 

\noindent \textbf{Reasoning-Driven Image Editing.} Recent work transcends attribute-level edits by integrating VLM reasoners for reasoning editing. ThinkGen~\cite{jiao2026thinkgen} relies on auxiliary instruction refinement and scenario-specific rules. Unified Thinker~\cite{zhou2026unified} jointly fine-tunes reasoning and generative modules, risking mutual interference. ThinkRL-Edit~\cite{li2026thinkrl} disentangles semantic reasoning from image generation and expands reasoning exploration beyond denoising. EditThinker~\cite{li2025editthinker} and ReasonEdit~\cite{yin2026reasonedit} employ multi-round reflections to improve editing accuracy, yet incurring substantial inference latency. Most closely related, DDA-Thinker adopts a reasoning-centric paradigm with dual rewards to independently optimize the reasoner, demonstrating the effectiveness of reasoning-oriented optimization. However, this isolation neglects cross-modal dependencies, inevitably inducing credit misalignment and gradient conflicts. To resolve this limitation, following the reasoning-centric and dual-reward paradigm, KnowThinker introduces BPPO, synergizing BRR and CPAF to enable coordinated multi-reward optimization across reasoning and generation.

\begin{figure*}[h]
    \centering
    \captionsetup{skip=2pt}
    \caption{
    Representative examples from 8 knowledge-intensive expert disciplines and 58 sub-categories in ExpertVerse.}
    \includegraphics[width=0.99\textwidth]{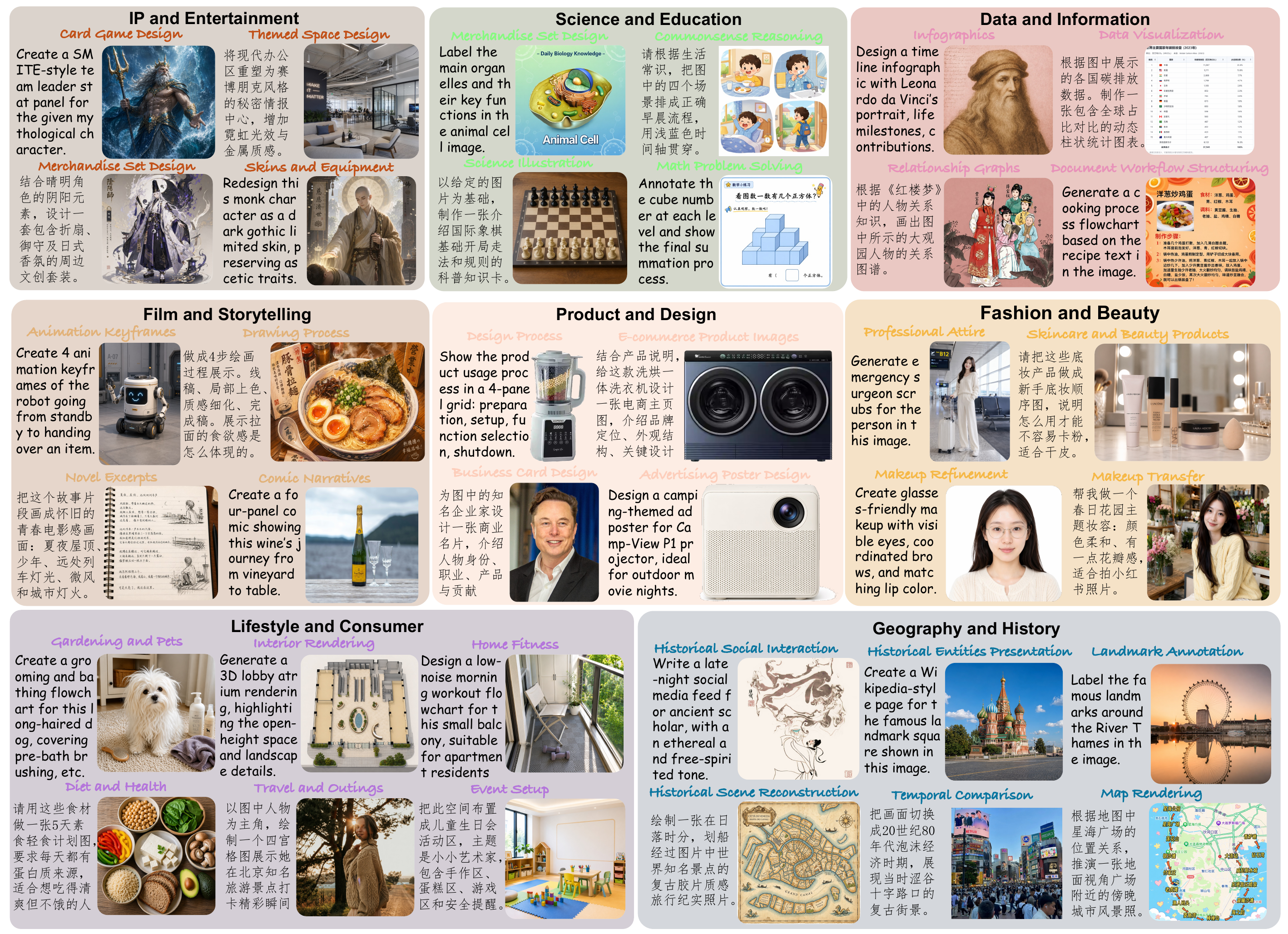}
    \label{fig:Overview_of_ExpertVerse}
\end{figure*}

\begin{figure}[]
    \centering
    \captionsetup{skip=5pt}
    \caption{
    Statistical overview of our ExpertVerse across task (\textit{left}) and capability dimensions (\textit{right}). Best viewed in color.}
    \includegraphics[width=0.44\textwidth]{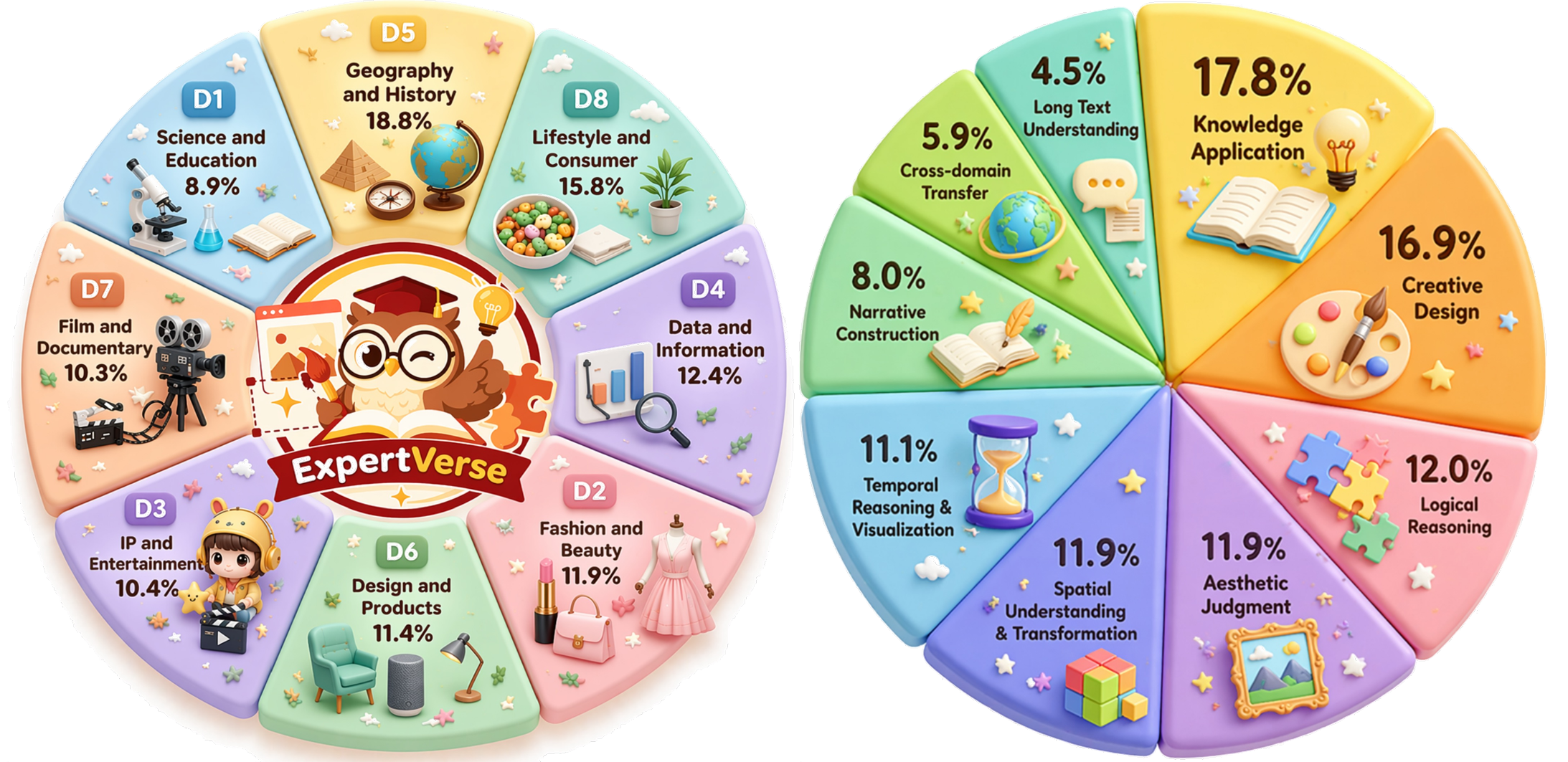}
    \label{fig:pie}
    \vspace{-0.49cm}
\end{figure}

\section{ExpertVerse}

\subsection{Preliminaries and Overview}
Given a textual prompt $p$ and an optional reference image $I_\text{ref}$, the objective of knowledge-intensive reasoning editing is to synthesize an output image $I_\text{out}$ that satisfies perceptual fidelity, visual aesthetics, and knowledge-based semantic constraints. Departing from reactive text-to-pixel mapping, this task enforces a reasoning-then-execution paradigm. Formally, the VLM-based reasoning policy (Thinker) externalizes an interpretable reasoning trace $R=\{r_1, r_2, \ldots, r_k\}$ and a refined instruction $q'$ via $(R, q')=\mathcal{T}(p, I_\text{ref}; \phi)$, grounded in latent world knowledge $\mathcal{K}$. Each step $r_i$ corresponds to expert-level cognitive operations, \textit{e.g.}, factual retrieval, causal inference, and visual planning. A visual execution policy (Generator $\mathcal{G}$) synthesizes the final output as $I_\text{out}\!=\!\mathcal{G}(q', I_\text{ref}, R)$. To this end, we develop \textbf{ExpertVerse}, a general knowledge-intensive reasoning editing benchmark covering a broad spectrum of knowledge domains and reasoning abilities. Distinct from prior studies, ExpertVerse features key superiorities: 
\begin{itemize}
    \item \textit{Broader scenarios and deeper expert knowledge.} ExpertVerse organizes reasoning tasks by 9 cognitive abilities and 8 expert disciplines, covering 58 sub-disciplines with high knowledge density and multi-hop knowledge reasoning.
    \item \textit{Richer task diversity and more diversified data.} ExpertVerse includes 1,611 expert-annotated instances with rationales and image captions across single-image editing, multi-image composition, and text-to-image synthesis. 
    \item \textit{Scalable data synthesis pipeline.} We propose a hierarchical expansion strategy that scales handcrafted prompts via VLMs, yielding ExpertVerse-100K, a large-scale dataset enriched with reasoning traces and refined instructions.
\end{itemize}

\subsection{Data Collection and Annotations}
To ensure comprehensive coverage and high knowledge density, we stratify reasoning-intensive editing tasks via an orthogonal taxonomy (Fig.~\ref{fig:pie}): \textit{9 cognitive capabilities} (knowledge application, creative design, logical reasoning, aesthetic judgment, spatial transformation, temporal reasoning, narrative construction, cross-domain transfer, and long text comprehension) and \textit{8 categories} (science\&education, fashion\&beauty, IP \&entertainment, data\&information, geography\&history, product\& design, film\&storytelling, and lifestyle\&consumer), yielding \textit{58 fine-grained sub-categories}. Reference images are primarily web-sourced, supplemented by a synthetic subset from proprietary T2I generation model. Guided by this taxonomy, we curate high-quality seed prompts—encompassing image descriptions, complex editing instructions, and knowledge-anchored rationales—as in-context exemplars to drive large-scale synthesis via proprietary VLM data collectors. To maximize diversity, we employ a \textit{hierarchical expansion strategy}, decomposing the 58 sub-categories into 3–6 atomic tasks each for uniform sample allocation. Finally, we enforce a dual-stage quality assurance pipeline: an automated VLM filter eliminates factual hallucinations and trivial or overly verbose instructions, followed by manual verification from three Ph.D. specialists to guarantee knowledge correctness and reasoning depth. To ensure rigorous evaluation, the benchmark is completely held-out and has never been exposed to the model during SFT, RL, reward modeling, or preliminary data distillation. ExpertVerse includes 1,611 instances spanning single-image editing (66.22\%), multi-image composition (7.5\%), and T2I synthesis (26.17\%), featuring Chinese/English annotations to evaluate cross-lingual instruction following.

We scale up the pipeline to curate ExpertVerse-100K, a large-scale training corpus that internalizes cross-disciplinary knowledge and planning capabilities via a proprietary VLM-generated reasoning CoT, refined instructions, and knowledge-anchored rationales. To the data acquisition bottleneck for high information-density charts in the \textit{Data \& Information} domain, we introduce a structured HTML-based rendering pipeline. We first leverage VLMs to generate semantic-rich HTML code for chart-heavy instances, which is rendered into high-fidelity webpage images. These HTML structures automatically yield paired images, tailored edit prompts.


\subsection{Evaluation Metrics}
To evaluate image generation models on ExpertVerse, we propose a four-dimensional evaluation protocol. Beyond widely adopted Visual Quality and Input Consistency, we introduce two novel metrics specifically tailored for knowledge-intensive synthesis: \textit{Knowledge Reasoning}, and \textit{Rationale Alignment}.

Among these, \textit{Knowledge Reasoning} evaluates the domain-knowledge grounding in generated images, assessing whether professional concepts, implicit constraints, and knowledge-dependent visual elements are accurately instantiated. \textit{Visual Quality} assesses perceptual fidelity, spatial composition, and stylistic coherence, alongside format-specific structural constraints, \textit{e.g.}, multi-panel topology and typographic legibility. \textit{Input Consistency} measures the preservation of subject identity and structural invariants (\textit{e.g.}, facial features and spatial layouts)—constraining valid deviations strictly to explicit instructional edits. \textit{Rationale Alignment} examines whether the output faithfully follows the guidance in reasoning rationale, including professional details, layout constraints, aesthetic and stylistic constraints, and intended narrative. Each metric is scored on a 1--5 scale and reported as a percentage in experiment results. Further, we utilize a separate proprietary multimodal evaluator as the evaluation model and design dimension-specific prompts to support consistent assessment. 

\section{KnowThinker: An Expert Knowledge Reasoner}
In this section, we utilize knowledge-aware supervision to train KnowThinker under thinker-centric paradigm, an MLLM reasoning engine that leverages internalized world knowledge to translate high-level user intent into executable visual plans. 

\subsection{Thinker-Centric Reasoning Training Paradigm}
Prior works~\cite{yin2026reasonedit,zhou2026unified} jointly optimize the VLM thinker and visual editors, which fundamentally suffers from the optimization incompatibility between discrete reasoning and continuous denoising policies. This misalignment severely complicates credit assignment and destabilizes training. In contrast, our \textit{Thinker-Centric} paradigm isolates the VLM thinker by freezing the visual editor as a deterministic rendering engine. Directing all optimization exclusively to the thinker enables it to focus purely on externalizing world knowledge into step-wise execution plans, thereby steering visual synthesis without the instability of joint policy updates.

To optimize the KnowThinker, we first adopt a SFT knowledge reasoning activation on a structured triplets including user prompts, thinking traces, and refined instructions from ExpertVerse-100K. This stage bootstraps KnowThinker's capacity to distill the inherent knowledge into visual plans. The training objective maximizes the log-likelihood of generating the reasoning trace $R$ and the refined instruction $q'$. 


\begin{table*}[h]
\centering
\vspace{-0.24cm}
\small
\setlength{\tabcolsep}{1.13pt}
\begin{tabular}{lcccccccccccccccccc}
\toprule
\multirow{2}{*}{\textbf{Models}} 
& \multicolumn{2}{c}{\textbf{Sci.\&Edu.}} 
& \multicolumn{2}{c}{\textbf{Fash.\&Beau.}} 
& \multicolumn{2}{c}{\textbf{IP\&Enter.}} 
& \multicolumn{2}{c}{\textbf{Data\&Info.}} 
& \multicolumn{2}{c}{\textbf{Geo.\&His.}} 
& \multicolumn{2}{c}{\textbf{Prod.\&Des.}} 
& \multicolumn{2}{c}{\textbf{Film\&Story}} 
& \multicolumn{2}{c}{\textbf{Life\&Consu.}} 
& \multirow{2}{*}{\textbf{Avg}} \\
\cmidrule(lr){2-3}
\cmidrule(lr){4-5}
\cmidrule(lr){6-7}
\cmidrule(lr){8-9}
\cmidrule(lr){10-11}
\cmidrule(lr){12-13}
\cmidrule(lr){14-15}
\cmidrule(lr){16-17}
& T2I & I2I
& T2I & I2I
& T2I & I2I
& T2I & I2I
& T2I & I2I
& T2I & I2I
& T2I & I2I
& T2I & I2I
& \\
\midrule
FLUX.1 Kontext 
& \cellcolor[HTML]{E7F7F9} 3.02 & \cellcolor[HTML]{E7F7F9} 12.06 
& \cellcolor[HTML]{E7F7F9} 53.13 & \cellcolor[HTML]{E7F7F9} 61.58 
& \cellcolor[HTML]{E7F7F9} 50.15 & \cellcolor[HTML]{E7F7F9} 54.63 
& \cellcolor[HTML]{E7F7F9} 5.43 & \cellcolor[HTML]{E7F7F9} 11.27 
& \cellcolor[HTML]{E7F7F9} 10.47 & \cellcolor[HTML]{E7F7F9} 16.84
& \cellcolor[HTML]{E7F7F9} 28.37 & \cellcolor[HTML]{E7F7F9} 31.79 
& \cellcolor[HTML]{E7F7F9} 21.41 & \cellcolor[HTML]{E7F7F9} 37.89 
& \cellcolor[HTML]{E7F7F9} 16.56 & \cellcolor[HTML]{E7F7F9} 23.14 
& \cellcolor[HTML]{E7F7F9} 28.67 \\

GPT-Image-1 
& \cellcolor[HTML]{E7F7F9} 22.38 & \cellcolor[HTML]{E7F7F9} 18.24 
& \cellcolor[HTML]{E7F7F9} 75.76 & \cellcolor[HTML]{E7F7F9} 54.79 
& \cellcolor[HTML]{E7F7F9} 57.86 & \cellcolor[HTML]{E7F7F9} 56.56 
& \cellcolor[HTML]{E7F7F9} 5.22 & \cellcolor[HTML]{E7F7F9} 8.66 
& \cellcolor[HTML]{E7F7F9} 33.89 & \cellcolor[HTML]{E7F7F9} 41.89 
& \cellcolor[HTML]{E7F7F9} 43.75 & \cellcolor[HTML]{E7F7F9} 36.50 
& \cellcolor[HTML]{E7F7F9} 61.33 & \cellcolor[HTML]{E7F7F9} 53.48 
& \cellcolor[HTML]{E7F7F9} 48.95 & \cellcolor[HTML]{E7F7F9} 34.97 
& \cellcolor[HTML]{E7F7F9} 41.52 \\

Nano Banana 
& \cellcolor[HTML]{E7F7F9} 29.32 & \cellcolor[HTML]{E7F7F9} 18.86 
& \cellcolor[HTML]{E7F7F9} 77.66 & \cellcolor[HTML]{E7F7F9} 45.83 
& \cellcolor[HTML]{E7F7F9} 87.38 & \cellcolor[HTML]{E7F7F9} 60.22 
& \cellcolor[HTML]{E7F7F9} 25.08 & \cellcolor[HTML]{E7F7F9} 22.91 
& \cellcolor[HTML]{E7F7F9} 49.99 & \cellcolor[HTML]{E7F7F9} 32.49 
& \cellcolor[HTML]{E7F7F9} 59.52 & \cellcolor[HTML]{E7F7F9} 39.73 
& \cellcolor[HTML]{E7F7F9} 64.74 & \cellcolor[HTML]{E7F7F9} 34.20 
& \cellcolor[HTML]{E7F7F9} 49.95 & \cellcolor[HTML]{E7F7F9} 40.34 
& \cellcolor[HTML]{E7F7F9} 46.33 \\

Nano Banana Pro 
& \cellcolor[HTML]{E7F7F9} 82.46 & \cellcolor[HTML]{E7F7F9} 58.80 
& \cellcolor[HTML]{E7F7F9} 89.98 & \cellcolor[HTML]{E7F7F9} 43.18 
& \cellcolor[HTML]{E7F7F9} 92.56 & \cellcolor[HTML]{E7F7F9} 58.83 
& \cellcolor[HTML]{E7F7F9} 74.72 & \cellcolor[HTML]{E7F7F9} 55.24 
& \cellcolor[HTML]{E7F7F9} 82.26 & \cellcolor[HTML]{E7F7F9} 50.90 
& \cellcolor[HTML]{E7F7F9} 86.86 & \cellcolor[HTML]{E7F7F9} 66.47 
& \cellcolor[HTML]{E7F7F9} 79.09 & \cellcolor[HTML]{E7F7F9} 49.99 
& \cellcolor[HTML]{E7F7F9} 91.21 & \cellcolor[HTML]{E7F7F9} 61.98 
& \cellcolor[HTML]{E7F7F9} 68.89 \\

Seedream-4.0
& \cellcolor[HTML]{E7F7F9} 40.70 & \cellcolor[HTML]{E7F7F9} 35.54 
& \cellcolor[HTML]{E7F7F9} 83.97 & \cellcolor[HTML]{E7F7F9} 65.41
& \cellcolor[HTML]{E7F7F9} 79.78 & \cellcolor[HTML]{E7F7F9} 68.59 
& \cellcolor[HTML]{E7F7F9} 44.29 & \cellcolor[HTML]{E7F7F9} 34.33 
& \cellcolor[HTML]{E7F7F9} 48.34 & \cellcolor[HTML]{E7F7F9} 39.42 
& \cellcolor[HTML]{E7F7F9} 83.19 & \cellcolor[HTML]{E7F7F9} 64.98
& \cellcolor[HTML]{E7F7F9} 79.66 & \cellcolor[HTML]{E7F7F9} 54.28 
& \cellcolor[HTML]{E7F7F9} 70.90 & \cellcolor[HTML]{E7F7F9} 54.10
& \cellcolor[HTML]{E7F7F9} 64.50 \\

Wan-2.7-Image-Pro 
& \cellcolor[HTML]{E7F7F9} 35.59 & \cellcolor[HTML]{E7F7F9} 23.35 
& \cellcolor[HTML]{E7F7F9} 80.60 & \cellcolor[HTML]{E7F7F9} 32.34 
& \cellcolor[HTML]{E7F7F9} 62.51 & \cellcolor[HTML]{E7F7F9} 38.04 
& \cellcolor[HTML]{E7F7F9} 31.50 & \cellcolor[HTML]{E7F7F9} 25.22 
& \cellcolor[HTML]{E7F7F9} 22.46 & \cellcolor[HTML]{E7F7F9} 14.25 
& \cellcolor[HTML]{E7F7F9} 67.42 & \cellcolor[HTML]{E7F7F9} 47.24   
& \cellcolor[HTML]{E7F7F9} 54.69 & \cellcolor[HTML]{E7F7F9} 34.62
& \cellcolor[HTML]{E7F7F9} 60.38 & \cellcolor[HTML]{E7F7F9} 36.58 
& \cellcolor[HTML]{E7F7F9} 40.79 \\

GPT-Image-2
& \cellcolor[HTML]{E2EBED} \textbf{94.34} & \cellcolor[HTML]{E2EBED} \textbf{62.04}
& \cellcolor[HTML]{E2EBED} \textbf{97.58} & \cellcolor[HTML]{E2EBED} \textbf{75.11} 
& \cellcolor[HTML]{E2EBED} \textbf{96.49} & \cellcolor[HTML]{E2EBED} \textbf{78.34} 
& \cellcolor[HTML]{E2EBED} \textbf{91.69} & \cellcolor[HTML]{E2EBED} \textbf{64.16} 
& \cellcolor[HTML]{E2EBED} \textbf{88.58} & \cellcolor[HTML]{E2EBED} \textbf{52.71} 
& \cellcolor[HTML]{E2EBED} \textbf{97.00} & \cellcolor[HTML]{E2EBED} \textbf{77.80}
& \cellcolor[HTML]{E2EBED} \textbf{92.44} & \cellcolor[HTML]{E2EBED} \textbf{67.14}
& \cellcolor[HTML]{E2EBED} \textbf{97.58} & \cellcolor[HTML]{E2EBED} \textbf{75.11} 
& \cellcolor[HTML]{E2EBED} \textbf{79.51} \\

\midrule
Echo-4o 
& \cellcolor[HTML]{FFFFE2}2.31 & \cellcolor[HTML]{FFFFE2}10.01 
& \cellcolor[HTML]{FFFFE2}41.25 & \cellcolor[HTML]{FFFFE2}43.00 
& \cellcolor[HTML]{FFFFE2}49.22 & \cellcolor[HTML]{FFFFE2}48.30 
& \cellcolor[HTML]{FFFFE2}3.42 & \cellcolor[HTML]{FFFFE2}7.04 
& \cellcolor[HTML]{FFFFE2}5.27 & \cellcolor[HTML]{FFFFE2}15.93 
& \cellcolor[HTML]{FFFFE2}18.20 & \cellcolor[HTML]{FFFFE2}21.47 
& \cellcolor[HTML]{FFFFE2}17.09 & \cellcolor[HTML]{FFFFE2}29.91 
& \cellcolor[HTML]{FFFFE2}11.49 & \cellcolor[HTML]{FFFFE2}14.87 
& \cellcolor[HTML]{FFFFE2}22.32 \\

Uni-CoT
& \cellcolor[HTML]{FFFFE2}3.10 & \cellcolor[HTML]{FFFFE2}9.98 
& \cellcolor[HTML]{FFFFE2}54.54 & \cellcolor[HTML]{FFFFE2}44.65 
& \cellcolor[HTML]{FFFFE2}56.35 & \cellcolor[HTML]{FFFFE2}49.32 
& \cellcolor[HTML]{FFFFE2}3.31 & \cellcolor[HTML]{FFFFE2}6.18 
& \cellcolor[HTML]{FFFFE2}17.54 & \cellcolor[HTML]{FFFFE2}19.83 
& \cellcolor[HTML]{FFFFE2}24.84 & \cellcolor[HTML]{FFFFE2}18.43 
& \cellcolor[HTML]{FFFFE2}28.83 & \cellcolor[HTML]{FFFFE2}23.67 
& \cellcolor[HTML]{FFFFE2}26.89 & \cellcolor[HTML]{FFFFE2}17.10 
& \cellcolor[HTML]{FFFFE2}26.45 \\

Step1X-Edit 
& \cellcolor[HTML]{FFFFE2}3.27 & \cellcolor[HTML]{FFFFE2}12.24 
& \cellcolor[HTML]{FFFFE2}33.22 & \cellcolor[HTML]{FFFFE2}46.98 
& \cellcolor[HTML]{FFFFE2}27.89 & \cellcolor[HTML]{FFFFE2}35.08 
& \cellcolor[HTML]{FFFFE2}3.42 & \cellcolor[HTML]{FFFFE2}4.02 
& \cellcolor[HTML]{FFFFE2}0.58 & \cellcolor[HTML]{FFFFE2}12.37 
& \cellcolor[HTML]{FFFFE2}10.17 & \cellcolor[HTML]{FFFFE2}16.52 
& \cellcolor[HTML]{FFFFE2}8.40 & \cellcolor[HTML]{FFFFE2}7.49 
& \cellcolor[HTML]{FFFFE2}3.61 & \cellcolor[HTML]{FFFFE2}12.96 
& \cellcolor[HTML]{FFFFE2}16.23 \\

Ovis-U1 
& \cellcolor[HTML]{FFFFE2}5.28 & \cellcolor[HTML]{FFFFE2}7.93 
& \cellcolor[HTML]{FFFFE2}36.90 & \cellcolor[HTML]{FFFFE2}43.26 
& \cellcolor[HTML]{FFFFE2}32.17 & \cellcolor[HTML]{FFFFE2}36.31 
& \cellcolor[HTML]{FFFFE2}2.42 & \cellcolor[HTML]{FFFFE2}4.93 
& \cellcolor[HTML]{FFFFE2}2.75 & \cellcolor[HTML]{FFFFE2}7.82 
& \cellcolor[HTML]{FFFFE2}9.56 & \cellcolor[HTML]{FFFFE2}6.39 
& \cellcolor[HTML]{FFFFE2}11.56 & \cellcolor[HTML]{FFFFE2}15.41 
& \cellcolor[HTML]{FFFFE2}4.61 & \cellcolor[HTML]{FFFFE2}6.16 
& \cellcolor[HTML]{FFFFE2}14.07 \\

BAGEL (w/o. CoT) 
& \cellcolor[HTML]{FFFFE2}3.19 & \cellcolor[HTML]{FFFFE2}8.47 
& \cellcolor[HTML]{FFFFE2}38.77 & \cellcolor[HTML]{FFFFE2}46.38 
& \cellcolor[HTML]{FFFFE2}29.88 & \cellcolor[HTML]{FFFFE2}47.06 
& \cellcolor[HTML]{FFFFE2}4.33 & \cellcolor[HTML]{FFFFE2}8.56
& \cellcolor[HTML]{FFFFE2}5.06 & \cellcolor[HTML]{FFFFE2}15.72
& \cellcolor[HTML]{FFFFE2}13.56 & \cellcolor[HTML]{FFFFE2}24.84
& \cellcolor[HTML]{FFFFE2}12.36 & \cellcolor[HTML]{FFFFE2}23.80 
& \cellcolor[HTML]{FFFFE2}10.02 & \cellcolor[HTML]{FFFFE2}17.40 
& \cellcolor[HTML]{FFFFE2}20.31 \\

DreamOmni2
& \cellcolor[HTML]{FFFFE2}15.78 & \cellcolor[HTML]{FFFFE2}23.01 
& \cellcolor[HTML]{FFFFE2}24.03 & \cellcolor[HTML]{FFFFE2}42.38 
& \cellcolor[HTML]{FFFFE2}21.03 & \cellcolor[HTML]{FFFFE2}34.59 
& \cellcolor[HTML]{FFFFE2}6.27 & \cellcolor[HTML]{FFFFE2}11.90 
& \cellcolor[HTML]{FFFFE2}18.88 & \cellcolor[HTML]{FFFFE2}23.53 
& \cellcolor[HTML]{FFFFE2}21.04 & \cellcolor[HTML]{FFFFE2}23.18 
& \cellcolor[HTML]{FFFFE2}15.15 & \cellcolor[HTML]{FFFFE2}34.26 
& \cellcolor[HTML]{FFFFE2}15.69 & \cellcolor[HTML]{FFFFE2}25.98 
& \cellcolor[HTML]{FFFFE2}29.28 \\

Qwen-Image (Edit) 
& \cellcolor[HTML]{FFFFE2}6.78 & \cellcolor[HTML]{FFFFE2}19.04
& \cellcolor[HTML]{FFFFE2}61.24 & \cellcolor[HTML]{FFFFE2}65.66
& \cellcolor[HTML]{FFFFE2}60.18 & \cellcolor[HTML]{FFFFE2}68.34 
& \cellcolor[HTML]{FFFFE2}8.03 & \cellcolor[HTML]{FFFFE2}12.21
& \cellcolor[HTML]{FFFFE2}24.02 & \cellcolor[HTML]{FFFFE2}37.98 
& \cellcolor[HTML]{FFFFE2}43.52 & \cellcolor[HTML]{FFFFE2}43.30 
& \cellcolor[HTML]{FFFFE2}28.61 & \cellcolor[HTML]{FFFFE2}53.43 
& \cellcolor[HTML]{FFFFE2}25.02 & \cellcolor[HTML]{FFFFE2}32.62 
& \cellcolor[HTML]{FFFFE2}38.81 \\


Know-Thinker (SFT) 
& \cellcolor[HTML]{FFFFE2}28.47 & \cellcolor[HTML]{FFFFE2}50.32 
& \cellcolor[HTML]{FFFFE2}76.99 & \cellcolor[HTML]{FFFFE2}82.18 
& \cellcolor[HTML]{FFFFE2}77.31 & \cellcolor[HTML]{FFFFE2}82.55 
& \cellcolor[HTML]{FFFFE2}20.71 & \cellcolor[HTML]{FFFFE2}44.06 
& \cellcolor[HTML]{FFFFE2}42.43 & \cellcolor[HTML]{FFFFE2}63.13 
& \cellcolor[HTML]{FFFFE2}66.29 & \cellcolor[HTML]{FFFFE2}70.73 
& \cellcolor[HTML]{FFFFE2}66.62 & \cellcolor[HTML]{FFFFE2}78.40
& \cellcolor[HTML]{FFFFE2}66.79 & \cellcolor[HTML]{FFFFE2}71.49 
& \cellcolor[HTML]{FFFFE2}64.22 \\

Know-Thinker (Ours)
& \cellcolor[HTML]{F4EAC9}\textbf{44.76} & \cellcolor[HTML]{F4EAC9}\textbf{51.05} 
& \cellcolor[HTML]{F4EAC9}\textbf{80.21} & \cellcolor[HTML]{F4EAC9}\textbf{92.38} 
& \cellcolor[HTML]{F4EAC9}\textbf{82.89} & \cellcolor[HTML]{F4EAC9}\textbf{85.87} 
& \cellcolor[HTML]{F4EAC9}\textbf{35.63} & \cellcolor[HTML]{F4EAC9}\textbf{50.61} 
& \cellcolor[HTML]{F4EAC9}\textbf{53.38} & \cellcolor[HTML]{F4EAC9}\textbf{66.06} 
& \cellcolor[HTML]{F4EAC9}\textbf{85.44} & \cellcolor[HTML]{F4EAC9}\textbf{71.06} 
& \cellcolor[HTML]{F4EAC9}\textbf{79.59} & \cellcolor[HTML]{F4EAC9}\textbf{82.49} 
& \cellcolor[HTML]{F4EAC9}\textbf{83.60} & \cellcolor[HTML]{F4EAC9}\textbf{75.47} 
& \cellcolor[HTML]{F4EAC9}\textbf{69.73} \\
\bottomrule
\end{tabular}
\caption{Results across diverse knowledge domains on ExpertVerse. KnowThinker outperforms all community models, achieving performance that closely approaches proprietary models. ``I2I'' includes single-image editing and multi-image composition.}
\label{tab:domain_performance}
\vspace{-0.4cm}
\end{table*}

We further improve the reasoning ability via RL with rule-based rewards, which refines knowledge-grounded reasoning via exploration. To mitigate inefficient exploration in direct RL, we employ a \textit{difficulty-aware data distillation} strategy to construct the RL training set $\mathcal{D}_{\text{RL}}$. Given the SFT-initialized policy $\pi_{\text{SFT}}$, we sample $N$ rollouts $\{(R_n, q'_n, I_n)\}_{n=1}^N$ for each instruction $p$. The reasoning and visual quality of each rollout are evaluated by specialized text and image reward model. To maximize contrastive signals for the advantage estimation, we filter instructions based on the group reward statistics: $\mathcal{D}_{\text{RL}}\!=\!\left\{p \middle|\tau_{\text{low}}\!\le\!\mu\!\le\tau_{\text{high}}, \sigma^2\!\ge\!\tau_{\text{var}} \right\}$. Retaining instances with a moderate mean (avoiding trivial cases) and a relatively high variance (ensuring the intra-group diversity) provides reliable supervision for subsequent RL optimization.

\begin{table*}[t]
\centering
\footnotesize
\setlength{\tabcolsep}{6.5pt}
\renewcommand{\arraystretch}{1}
\begin{tabular}{lcccccccc}
\toprule
\textbf{Model} 
& \textbf{Reasoning.}
& \textbf{Consistency.}
& \textbf{Visual.}
& \textbf{Temporal} 
& \textbf{Causal} 
& \textbf{Spatial} 
& \textbf{Logical} 
& \textbf{Overall} \\
\midrule

\rowcolor[HTML]{E7F7F9}
\multicolumn{9}{c}{\textit{Proprietary Models}} \\
Seedream-4.0
& 58.9 & 67.4 & 91.2 & 12.9 & 12.2 & 11.0 & 7.1 & 10.8 \\
GPT-Image-1-mini 
& 54.1 & 71.5 & 93.7 & 24.7 & 28.9 & 33.0 & 9.4 & 24.4 \\
GPT-Image-1
& 62.8 & 80.2 & 94.9 & 34.1 & 32.2 & 37.0 & 10.6 & 28.9 \\
NanoBanana
& 61.2 & 86.0 & 91.3 & 25.9 & 47.8 & 37.0 & 18.8 & 32.8 \\
NanoBanana-Pro
& 77.0 & 85.5 & 94.4 & 41.2 & 61.1 & 48.0 & 37.6 & 47.2 \\
GPT-Image-2 
& 73.8 & 89.3 & 94.9 & 45.9 & 66.7 & 50.0 & 34.1 & 49.4 \\
GPT-Image-1.5 
& 69.7 & 92.5 & 94.9 & 57.6 & 62.2 & 62.0 & 21.2 & 51.4 \\
\midrule

\rowcolor[HTML]{FFFFE2}
\multicolumn{9}{c}{\textit{Community Models}} \\
EMU2~\cite{sun2024generative} 
& 22.6 & 38.2 & 78.3 & 1.2 & 1.1 & 0.0 & 0.0 & 0.5 \\
OmniGen~\cite{xiao2025omnigen} 
& 25.1 & 41.5 & 73.5 & 1.2 & 1.0 & 0.0 & 1.2 & 0.8 \\
Step1X-Edit~\cite{liu2025step1x} 
& 30.3 & 12.6 & 74.9 & 0.0 & 2.2 & 2.0 & 3.5 & 1.9 \\
Ovis-U1~\cite{wang2025ovis} 
& 33.9 & 52.7 & 72.9 & 1.2 & 3.3 & 4.0 & 2.4 & 2.8 \\
FLUX.1-Dev~\cite{batifol2025flux} 
& 26.0 & 71.6 & 85.2 & 2.3 & 5.5 & 13.0 & 1.2 & 5.8 \\
BAGEL~\cite{deng2025emerging} 
& 45.9 & 73.8 & 80.1 & 5.9 & 17.8 & 21.0 & 1.2 & 11.9 \\
Uni-CoT~\cite{qin2025uni} 
& -- & -- & -- & 8.2 & 18.9 & 20.0 & 1.2 & 12.5 \\
ThinkGen~\cite{jiao2026thinkgen} 
& -- & -- & -- & 16.4 & 17.7 & 16.0 & 1.1 & 13.0 \\
DeepGen~\cite{wang2026deepgen} 
& -- & -- & -- & 15.3 & 18.9 & 14.0 & 4.7 & 13.3 \\
EditThinker~\cite{li2025editthinker} 
& -- & -- & -- & 10.8 & 23.3 & 27.0 & 8.2 & 17.8 \\
Unified Thinker~\cite{zhou2026unified} 
& 61.9 & 76.2 & 90.5 & 32.9 & 30.0 & 41.0 & 9.4 & 28.9 \\
ThinkRL-Edit~\cite{li2026thinkrl} 
& 61.7 & 81.6 & 62.5 & 18.8 & 37.5 & 25.0 & 37.5 & 29.7 \\
\midrule

\rowcolor[HTML]{EEE0FA}
Qwen-Image-Edit-2511 
& 49.9 & 71.0 & 91.5 & 21.2 & 18.9 & 31.0 & 4.7 & 19.4 \\
\quad + \tiny{Know-Thinker(Qwen3-VL-8B)} 
& 66.0 & 81.4 & 91.7 & 34.1 & 42.2 & 38.0 & 17.6 & 33.3 \\
\quad + \tiny{Know-Thinker(Qwen3.5-27B)} 
& \textbf{68.2} & \textbf{82.6} & \textbf{93.7} 
& \textbf{44.7} & \textbf{52.2} & \textbf{46.0} 
& \textbf{15.3} & \textbf{40.0} \\
\bottomrule
\end{tabular}
\vspace{-0.2cm}
\caption{Quantitative result comparisons with proprietary and community image editing models on RISEBench benchmark. We report three general performance metrics: Instruction Reasoning (Reasoning.), Appearance Consistency (Consistency.), and Visual Plausibility (Visual.). Additionally, we present category-wise accuracy (\%) for four specific reasoning dimensions: Temporal, Causal,
Spatial, and Logical. The Overall score is the average of these four category-wise accuracies.}
\label{tab:main_results}
\vspace{-0.4cm}
\end{table*}

\vspace{-0.2cm}
\subsection{Bootstrapped Pareto Policy Optimization}
\vspace{-0.1cm}
Unlike token-level imitation in SFT, current methods employ GRPO to facilitate trajectory-level exploration and reward-driven policy optimization. However, the GRPO suffers from cross-modal credit misalignment and multi-objective gradient conflict when optimizing over heterogeneous rewards. We thus propose a novel BPPO algorithm for reasoning edit tasks. In knowledge-intensive generation, this independence leads to severe modality imbalance, where a rollout exhibit coherent reasoning but poor visual fidelity (or vice versa), yet receive a positive advantage in its stronger domain. 

\noindent
\textbf{Bootstrapping Reward Rectification (BRR)}. Towards the cross-modal credit misalignment, we propose BRR, a baseline-level calibration that rectifies domain-wise advantages with paired cross-modal feedback. Unlike naive reward weighting, BRR calibrates rewards before advantage estimation, allowing heterogeneous domains to regularize each other and mitigating biased policy updates from domain-specific rewards. 


Let $\mathcal{M}\!=\!\{\text{cot}, \text{img}\}$ denote the paired refined instruction and generated image. $r_i^{m}$ is reward in domain $m\!\in\!\mathcal{M}$, with group mean $\mu^{m}$ and standard deviation $\sigma^{m}$. The standard intra-domain advantage in GRPO is defined as $A_i^{m}\!=\!(r_i^{m}\!-\!\mu^{m})/\sigma^{m}$.
Instead of relying solely on the intra-domain baseline $\mu^{m}$, we introduce a cross-modal anchor $s_i^{\bar{m}}$ derived from the paired domain $\bar{m} \in \mathcal{M}\!\setminus\!\{m\}$. We dynamically rectify the baseline by bootstrapping the group mean with the cross-modal anchor:
\begin{equation}
\setlength{\abovedisplayskip}{2pt}
\setlength{\belowdisplayskip}{2pt}
    \tilde{b}_i^{m} = (1 - \beta) \mu^{m} + \beta s_i^{\bar{m}},
\end{equation}
where $\beta\!\in\![0, 1]$ modulates the regularization intensity of cross-modal reward signals. Substituting the rectified baseline $\tilde{b}_i^{m}$ into advantage formula derives the calibrated advantage:
\vspace{-0.15cm}
\begin{equation}
\label{BRR}
\setlength{\abovedisplayskip}{5pt}
\setlength{\belowdisplayskip}{3pt}
    \tilde{A}_i^{m}\!=\!(r_i^{m}\!-\!\tilde{b}_i^{m})/{\sigma^{m}} = A_i^{m} - \beta\left( s_i^{\bar{m}} - \mu^{\bar{m}} \right)/\sigma^{m}.
\end{equation}
\noindent
Eq.~\ref{BRR} rectifies the reward in domain $m$ with an adjustment of $-\beta (s_i^{\bar{m}}\!-\!\mu^{\bar{m}})$. The negative sign penalizes domain imbalance by elevating the baseline for domain $m$ when its paired domain excels ($s_i^{\bar{m}}\!>\!\mu^{\bar{m}}$), suppressing $A_i^{m}$ and preventing model overfitting on ``\textit{reasoning-strong but visually-weak}'' rollouts.

\noindent
\textbf{Conflict-Aware Pareto Advantage Fusion (CPAF)}. GRPO aggregates multi-objective rewards via the naive linear combination, suppressing weaker objectives and inducing gradient conflicts. Moving beyond reward-level aggregation, we operate directly in the advantage space. We derive group-whitened advantages $A_i^{\text{img}}$ and $A_i^{\text{cot}}$ for the generated image and refined instruction, normalized within each rollout group $i$, then fuse them via sign-based conflict detection as $\hat{A}_i$:
\begin{equation}
\setlength{\abovedisplayskip}{1pt}
\setlength{\belowdisplayskip}{1pt}
\label{CPAF}
\hspace*{-0.2cm}
    \hat{A}_i\!=\!\!
    \begin{cases}
        \alpha A_i^{\text{img}}\!+\!(1-\alpha) A_i^{\text{cot}},\!\! & \text{if } \text{sgn}(A_i^{\text{img}}) = \text{sgn}(A_i^{\text{cot}}), \\
        \mathcal{C}(A_i^{\text{img}}, A_i^{\text{cot}}),\! & \text{otherwise}.
    \end{cases}
\end{equation}
where $\alpha\!\in\![0,1]$ is fusion weight. In responses ($\text{sgn}(A_i^{\text{img}})\!\neq\!\text{sgn}(A_i^{\text{cot}})$), these two domains yield antagonistic gradients. Weighted summation in such responses forces a trade-off that inevitably sacrifices one objective, breaking pareto optimality. To preserve pareto consistency, we adopt a min-operator:
\begin{equation}
\setlength{\abovedisplayskip}{2pt}
\setlength{\belowdisplayskip}{2pt}
    \mathcal{C}(A_i^{\text{img}}, A_i^{\text{cot}}) = \operatorname{argmin}_{A \in \{A_i^{\text{img}}, A_i^{\text{cot}}\}} |A|.
\end{equation}
\noindent
By retreating along the weaker signal, $\mathcal{C}$ implicitly bounds the update step in conflict regions, ensuring stable and pareto-improving policy refinement across heterogeneous objectives.

\begin{table*}[h]
\centering
\small
\setlength{\tabcolsep}{2.55pt}
\begin{tabular}{lccccccccccc}
\toprule
\multirow{2}{*}{\textbf{Model}} 
& \multicolumn{4}{c}{\textbf{Factual Knowledge}} 
& \multicolumn{3}{c}{\textbf{Conceptual Knowledge}} 
& \multicolumn{3}{c}{\textbf{Procedural Knowledge}} 
& \multirow{2}{*}{\shortstack{\textbf{Overall}\\\textbf{Average}}} \\
\cmidrule(lr){2-5} \cmidrule(lr){6-8} \cmidrule(lr){9-11}
& \textbf{Attribute} & \textbf{Spatial} & \textbf{Temporal} & \textbf{Avg.}
& \textbf{Social} & \textbf{Natural} & \textbf{Avg.}
& \textbf{Logical} & \textbf{Instruction} & \textbf{Avg.}
&  \\
\midrule

\rowcolor{yellow!20}
\multicolumn{12}{c}{\textit{Proprietary Models}} \\
\midrule
Gemini 2.0 
& 66.33 & 63.33 & 63.92 & 65.26 
& 68.19 & 56.94 & 59.65 
& 54.13 & 71.67 & 62.90 
& 62.41 \\

NanoBanana 
& 78.97 & 80.17 & 86.49 & 80.48 
& 83.70 & 76.34 & 78.11 
& 63.96 & 90.44 & 75.31 
& 78.17 \\

GPT-4o 
& 83.17 & 79.08 & 68.25 & 79.80 
& 85.50 & 80.06 & 81.37 
& 71.56 & 85.08 & 78.32 
& 80.09 \\

NanoBanana-Pro 
& 85.61 & 87.17 & 88.06 & 86.36 
& 85.25 & 83.64 & 84.03 
& 81.42 & 93.22 & 86.48 
& 85.31 \\

\midrule
\rowcolor[HTML]{92FFCC}
\multicolumn{12}{c}{\textit{Community Models}} \\
\midrule
InstructPix2Pix~\cite{brooks2023instructpix2pix}
& 30.33 & 21.33 & -- & 23.33 
& 22.56 & 26.56 & 25.59 
& 19.81 & 14.75 & 17.28 
& 22.82 \\

OmniGen~\cite{xiao2025omnigen}
& 37.92 & 28.25 & 21.83 & 33.11 
& 30.63 & 27.19 & 28.02 
& 11.94 & 35.83 & 23.89 
& 28.85 \\

MagicBrush~\cite{zhang2023magicbrush} 
& 53.92 & 39.58 & -- & 41.84 
& 42.94 & 38.06 & 39.24 
& 30.00 & 23.08 & 26.54 
& 37.15 \\

AnyEdit~\cite{jiang2025anyedit}
& 47.67 & 45.17 & -- & 39.26 
& 38.56 & 42.94 & 41.88 
& 36.56 & 26.92 & 31.74 
& 38.55 \\

Emu2~\cite{sun2024generative}
& 51.50 & 48.83 & 22.17 & 45.40 
& 34.69 & 38.44 & 37.54 
& 24.81 & 45.00 & 34.91 
& 39.70 \\

Step1X-Edit~\cite{liu2025step1x}
& 55.50 & 51.75 & -- & 45.52 
& 44.69 & 49.06 & 48.01 
& 40.88 & 22.75 & 31.82 
& 43.29 \\

HiDream-E1~\cite{cai2025hidream}
& 52.75 & 49.42 & -- & 43.31 
& 52.56 & 49.25 & 50.05 
& 45.19 & 30.08 & 37.64 
& 44.72 \\

ByteMorph~\cite{chang2025bytemorph} 
& 61.17 & 62.00 & -- & 51.27 
& 45.50 & 47.38 & 46.92 
& 32.00 & 31.33 & 31.67 
& 44.85 \\

FLUX.1-Dev~\cite{batifol2025flux} 
& 64.83 & 60.92 & -- & 53.28 
& 48.94 & 50.81 & 50.36 
& 46.06 & 39.00 & 42.53 
& 49.54 \\

OmniGen2~\cite{wu2025omnigen2} 
& 59.92 & 52.25 & 54.75 & 57.36 
& 47.56 & 43.12 & 44.20 
& 32.50 & 63.08 & 47.79 
& 49.71 \\

UniWorld-V1~\cite{lin2025uniworld}
& 58.17 & 54.50 & 63.00 & 47.71 
& 47.50 & 43.94 & 44.80 
& 42.00 & 53.83 & 47.92 
& 50.27 \\

Step1X-Edit v1.1~\cite{liu2025step1x} 
& 64.17 & 61.75 & -- & 53.05 
& 52.06 & 55.06 & 54.34 
& 52.56 & 36.75 & 44.66 
& 51.59 \\

BAGEL~\cite{deng2025emerging}
& 64.27 & 62.42 & 42.45 & 60.26 
& 55.40 & 56.01 & 55.86 
& 52.54 & 50.56 & 51.69 
& 56.21 \\

BAGEL-Think~\cite{deng2025emerging}
& 67.42 & 68.33 & 58.67 & 66.18 
& 63.55 & 61.40 & 61.92 
& 48.12 & 50.22 & 49.02 
& 60.18 \\

Uni-CoT~\cite{qin2025uni}
& 72.76 & 72.87 & 67.10 & 71.85 
& 70.81 & 66.00 & 67.16 
& 53.43 & 73.93 & 63.68 
& 68.00 \\

ThinkRL-Edit~\cite{li2026thinkrl}
& 81.02 & 81.45 & -- & 81.13 
& 75.67 & 71.25 & 72.31 
& 49.07 & 79.71 & 57.44 
& 71.65 \\

EditThinker~\cite{li2025editthinker} 
& 78.48 & 73.83 & -- & 77.24 
& 76.20 & 70.69 & 72.02 
& 65.23 & 66.89 & 65.94 
& 71.91 \\

\midrule
\rowcolor[HTML]{EEE0FA}
Qwen-Image-Edit-2511 
& 74.85 & 84.50 & 68.13 & 75.89  
& 66.10 & 59.54 & 61.15 
& 58.42 & 79.61 & 67.55 
& 67.14 \\

\quad + \tiny{Know-Thinker(Qwen3-VL-8B)} 
& 78.13 & 82.54 & 80.92 & 80.53
& 80.18 & 78.75 & 79.47
& 69.03 & 85.48 & 77.26 
& 79.09 \\

\quad + \tiny{Know-Thinker(Qwen3.5-27B)}  
& 79.78 & 86.93 & 82.53 & 83.08
& 84.72 & 80.57 & 82.65
& 74.77 & 91.87 & 83.32
& 83.02 \\

\bottomrule
\end{tabular}
\caption{Quantitative performance evaluation on the KRIS-Bench benchmark. Models are comprehensively assessed across three foundational knowledge types: \textbf{Factual Knowledge} (Attribute, Spatial, Temporal), \textbf{Conceptual Knowledge} (Social, Natural), and \textbf{Procedural Knowledge} (Logical, Instruction), alongside the \textbf{Overall Average} score.}
\label{tab:kris_bench_results}
\end{table*}

\begin{table*}[]
\centering
\footnotesize
\setlength{\tabcolsep}{1.15pt}
\begin{tabular}{lcccccccccccccccc}
\toprule
\multirow{2}{*}{Model} 
& \multicolumn{5}{c}{Awareness Task} 
& \multicolumn{5}{c}{Interpretation Task} 
& \multicolumn{5}{c}{Imagination Task} 
& \multirow{2}{*}{Overall} \\
\cmidrule(lr){2-6} \cmidrule(lr){7-11} \cmidrule(lr){12-16}
& $\mathcal{IF}\uparrow$ & $\mathcal{DP}\uparrow$ & $\mathcal{VQ}\uparrow$ & $\mathcal{KF}\uparrow$ & AVG
& $\mathcal{IF}\uparrow$ & $\mathcal{DP}\uparrow$ & $\mathcal{VQ}\uparrow$ & $\mathcal{KF}\uparrow$ & AVG
& $\mathcal{IF}\uparrow$ & $\mathcal{DP}\uparrow$ & $\mathcal{VQ}\uparrow$ & $\mathcal{CF}\uparrow$ & AVG
& AVG \\
\midrule
Nano Banana~\cite{comanici2025gemini} 
& 70.6 & 85.7 & 86.8 & 75.2 & 79.6 
& 63.4 & 84.9 & 91.4 & 61.5 & 75.3 
& 75.3 & 73.8 & 87.3 & 44.3 & 70.2 
& 75.0 \\

Seedream 4.0~\cite{seedream2025seedream} 
& 70.8 & 78.1 & 86.6 & 74.6 & 77.5 
& 63.7 & 80.1 & 90.6 & 64.2 & 74.6 
& 82.2 & 77.8 & 86.9 & 47.0 & 73.5 
& 75.2 \\

GPT-image-1~\cite{hurst2024gpt} 
& 78.5 & 85.8 & \textbf{88.0} & 81.2 & 83.3 
& 62.9 & 82.9 & \textbf{93.0} & 60.8 & 74.9 
& 84.4 & 76.2 & \textbf{89.2} & 48.4 & 74.6 
& 77.6 \\

Nano Banana Pro~\cite{nanobananapro} 
& \textbf{85.4} & \textbf{88.6} & 83.9 & \textbf{91.4} & \textbf{87.3} 
& \textbf{76.0} & \textbf{89.1} & 92.3 & \textbf{75.8} & \textbf{83.3} 
& \textbf{86.6} & \textbf{79.5} & 88.8 & \textbf{51.5} & \textbf{76.6} 
& \textbf{82.4} \\
\midrule

MagicBrush~\cite{zhang2023magicbrush} 
& 27.2 & 43.4 & 53.3 & 27.1 & 37.8 
& 16.8 & 50.3 & 63.2 & 22.2 & 38.1 
& 18.0 & 36.9 & 44.8 & 22.3 & 30.5 
& 35.5 \\

OmniGen~\cite{xiao2025omnigen} 
& 35.0 & 42.0 & 46.7 & 37.4 & 40.3 
& 19.0 & 34.8 & 40.3 & 21.5 & 28.9 
& 42.2 & 35.1 & 46.0 & 38.7 & 40.5 
& 36.6 \\

AnyEdit~\cite{jiang2025anyedit}
& 25.0 & 54.6 & 61.3 & 26.3 & 41.8 
& 15.9 & 61.2 & 62.0 & 20.2 & 39.8 
& 9.1 & 49.7 & 50.9 & 16.5 & 31.5 
& 37.7 \\

UltraEdit~\cite{zhao2024ultraedit}
& 26.5 & 42.5 & 53.1 & 33.9 & 39.0 
& 24.3 & 61.7 & 73.6 & 26.7 & 46.6 
& 20.7 & 31.7 & 45.8 & 27.5 & 31.5 
& 39.0 \\

ICEdit~\cite{zhang2025icedit} 
& 26.1 & 42.2 & 61.2 & 31.8 & 40.4 
& 21.4 & 48.3 & 81.5 & 24.9 & 44.0 
& 21.5 & 40.6 & 54.0 & 25.0 & 35.3 
& 39.9 \\

FLUX.1-Dev~\cite{batifol2025flux}  
& 31.4 & 52.0 & 55.0 & 35.5 & 43.5 
& 27.5 & 62.2 & 69.6 & 29.0 & 47.1 
& 39.1 & 47.1 & 43.4 & 27.1 & 39.2 
& 43.2 \\

FLUX.2-Dev~\cite{flux-2-2025} 
& 42.6 & 63.3 & 78.4 & 53.3 & 59.4 
& 35.4 & 75.0 & 85.6 & 37.6 & 58.4 
& 73.6 & 70.7 & 82.1 & 43.6 & 67.5 
& 61.8 \\
\midrule

UniWorld-V1~\cite{lin2025uniworld} 
& 31.5 & 48.9 & 58.8 & 38.6 & 44.5 
& 18.1 & 44.5 & 58.1 & 22.5 & 35.8 
& 30.3 & 50.3 & 64.2 & 27.5 & 43.1 
& 41.1 \\

HiDream-E1~\cite{cai2025hidream} 
& 29.7 & 41.2 & 56.3 & 32.0 & 39.8 
& 26.7 & 53.6 & 68.4 & 29.6 & 44.6 
& 39.6 & 40.1 & 49.9 & 29.6 & 39.8 
& 41.4 \\

OmniGen2~\cite{wu2025omnigen2} 
& 35.0 & 64.0 & 75.4 & 41.3 & 53.9 
& 18.9 & 56.9 & 64.9 & 23.5 & 41.1 
& 42.0 & 64.4 & 74.6 & 31.8 & 53.2 
& 49.4 \\

Step1X-Edit-v1p2~\cite{liu2025step1x} 
& 39.8 & 53.5 & 61.3 & 44.4 & 49.7 
& 35.7 & 73.0 & 75.2 & 38.2 & 55.5 
& 44.7 & 49.4 & 50.3 & 28.4 & 43.2 
& 49.5 \\

Echo-4o~\cite{ye2025echo} 
& 47.6 & 63.0 & 75.4 & 51.7 & 59.4 
& 30.8 & 71.0 & 80.4 & 32.9 & 53.8 
& 63.4 & 62.4 & 73.7 & 41.2 & 60.2 
& 57.8 \\

Bagel~\cite{sun2024generative} 
& 46.2 & 71.0 & 75.8 & 50.8 & 61.0 
& 38.6 & 72.1 & 78.8 & 39.5 & 57.3 
& 62.8 & 68.5 & 74.5 & 40.7 & 61.6 
& 60.0 \\

Uni-CoT~\cite{qin2025uni} 
& 46.0 & 69.1 & 77.8 & 51.6 & 61.1 
& 36.9 & 70.1 & 76.3 & 38.6 & 55.5 
& 67.6 & 64.3 & 79.6 & 42.9 & 63.6 
& 60.1 \\

Qwen-Image-Edit~\cite{wu2025qwen} 
& 48.1 & 69.0 & 79.5 & 53.6 & 62.5 
& 32.1 & 69.7 & 80.6 & 34.2 & 54.1 
& 67.1 & 66.8 & 79.2 & 42.3 & 63.8 
& 60.2 \\

DreamOmni2~\cite{cai2025hidream} 
& 43.3 & 74.4 & 85.0 & 51.2 & 63.5 
& 34.3 & 81.7 & 88.1 & 35.9 & 60.0 
& 50.6 & 64.9 & 81.9 & 35.3 & 58.2 
& 60.6 \\
\midrule
Know-Thinker (Ours)
& \textbf{73.3} & \textbf{81.8} & \textbf{85.9} & \textbf{73.6} & \textbf{78.7} 
& \textbf{59.2} & \textbf{83.9} & \textbf{91.3} & \textbf{57.6} & \textbf{73.0} 
& \textbf{88.0} & \textbf{84.2} & \textbf{92.7} & \textbf{56.1} & \textbf{80.3} 
& \textbf{77.3}
\\
\bottomrule
\end{tabular}
\caption{Main results on the English version of WiseEdit~\cite{pan2026wiseedit} benchmark. The best results are marked in bold for open- and closed-models, respectively. Notably, our KnowThinker establishes a new state-of-the-art among open-source models.}
\label{tab:wiseedit_main}
\end{table*}

\begin{table*}[t]
\centering
\resizebox{\textwidth}{!}{
\begin{tabular}{lcccccc cccccc c}
\toprule
\multirow{2}{*}{Model}
& \multicolumn{6}{c}{\textit{English Version}}
& \multicolumn{6}{c}{\textit{Chinese Version}}
& \multicolumn{1}{c}{Overall} \\
\cmidrule(lr){2-7} \cmidrule(lr){8-13} \cmidrule(lr){14-14}
& $\mathcal{IF}\uparrow$ & $\mathcal{DP}\uparrow$ & $\mathcal{VQ}\uparrow$ & $\mathcal{KF}\uparrow$ & $\mathcal{CF}\uparrow$ & AVG
& $\mathcal{IF}\uparrow$ & $\mathcal{DP}\uparrow$ & $\mathcal{VQ}\uparrow$ & $\mathcal{KF}\uparrow$ & $\mathcal{CF}\uparrow$ & AVG
& AVG \\
\midrule
AnyEdit~\cite{jiang2025anyedit}
& 2.5 & 5.6 & 20.6 & 3.3 & 11.7 & 8.7
& 1.3 & 5.1 & 22.2 & 2.9 & 9.3 & 8.2
& 8.5 \\

OmniGen~\cite{xiao2025omnigen}
& 23.5 & 25.7 & 41.2 & 31.5 & 48.9 & 34.2
& 4.4 & 15.4 & 50.3 & 15.1 & 32.2 & 23.5
& 28.9 \\

FLUX.2 Dev~\cite{flux-2-2025}
& 42.3 & 68.5 & 75.6 & 56.5 & 49.1 & 58.4
& 46.3 & 73.4 & 80.3 & 59.9 & 52.8 & 62.6
& 60.5 \\
\midrule

UniWorld-V1~\cite{lin2025uniworld}
& 18.1 & 32.1 & 55.3 & 22.8 & 28.6 & 31.4
& 8.8 & 23.8 & 64.6 & 12.4 & 15.4 & 25.0
& 28.2 \\

OmniGen2~\cite{wu2025omnigen2}
& 34.1 & 50.2 & 72.4 & 49.0 & 44.8 & 50.1
& 30.2 & 51.7 & 75.1 & 47.3 & 45.8 & 50.0
& 50.1 \\

DreamOmni2~\cite{cai2025hidream}
& 34.8 & 62.3 & 78.8 & 46.0 & 41.0 & 52.6
& 36.6 & 51.6 & 79.1 & 49.6 & 35.5 & 50.4
& 51.5 \\

Echo-4o~\cite{ye2025echo}
& 42.7 & 46.5 & 64.1 & 50.9 & 48.7 & 50.6
& 41.0 & 55.5 & 68.4 & 53.0 & 50.1 & 53.6
& 52.1 \\

Qwen-Image-Edit~\cite{wu2025qwen}
& 38.7 & 58.6 & 75.8 & 48.5 & 47.1 & 53.8
& 35.3 & 55.0 & 78.1 & 49.6 & 48.9 & 53.4
& 53.6 \\

Bagel~\cite{sun2024generative}
& 43.8 & 62.0 & 70.0 & 53.5 & 43.4 & 54.5
& 39.5 & 58.2 & 73.5 & 55.5 & 46.8 & 54.7
& 54.6 \\

Uni-CoT~\cite{qin2025uni}
& 35.5 & 60.0 & 69.3 & 57.0 & 49.5 & 54.3
& 36.4 & 55.2 & 77.4 & 58.1 & 48.9 & 55.2
& 54.8 \\
\midrule

Nano Banana~\cite{comanici2025gemini}
& 53.8 & 75.2 & 82.7 & 82.4 & 53.7 & 69.6
& 53.3 & 71.3 & 79.9 & 77.4 & 51.1 & 66.6
& 68.1 \\

GPT-image-1~\cite{hurst2024gpt}
& 58.7 & 75.9 & 87.6 & 77.9 & 54.8 & 71.0
& 59.5 & 76.6 & 88.2 & 78.8 & 54.1 & 71.4
& 71.2 \\

Seedream 4.0~\cite{seedream2025seedream}
& 67.2 & 77.3 & 79.6 & 89.7 & 53.1 & 73.4
& 59.3 & 62.8 & 81.0 & 87.8 & 55.5 & 69.3
& 71.4 \\

Nano Banana Pro~\cite{nanobananapro}
& 68.1 & 78.1 & 86.7 & 88.1 & 56.6 & 75.5
& 77.7 & \textbf{83.8} & 84.3 & \textbf{90.7} & 57.0 & 78.7
& 77.1 \\
\midrule
KnowThinker (Ours)
& \textbf{89.3} & \textbf{83.7} & \textbf{86.8} & \textbf{89.7} & \textbf{59.7} & \textbf{81.8}
& \textbf{87.2} & 82.8 & \textbf{94.2} & 87.8 & \textbf{62.4} & \textbf{82.9}
& \textbf{82.4} \\
\bottomrule
\end{tabular}
}
\vspace{-0.15cm}
\caption{Main results on WiseEdit-Complex~\cite{pan2026wiseedit}. We exclude models unable to handle multi-image inputs.}
\label{tab:WiseEdit-Complex}
\vspace{-0.2cm}
\end{table*}

\vspace{-0.4cm}
\begin{table*}[t]
\centering
\footnotesize
\setlength{\tabcolsep}{1.35pt}
\begin{tabular}{lcccc ccccc cc}
\toprule
\multirow{3}{*}{Model} & \multicolumn{5}{c}{Real World Scenario} & \multicolumn{5}{c}{Game World Scenario} & \multirow{3}{*}{Overall} \\
\cmidrule(lr){2-6} \cmidrule(lr){7-11}
& \multicolumn{1}{c}{\multirow{2}{*}{\shortstack{Attribute\\Modification}}} & \multicolumn{1}{c}{\multirow{2}{*}{\shortstack{Structure\\Transform}}} & \multicolumn{1}{c}{\multirow{2}{*}{\shortstack{Physical\\Interaction}}} & \multicolumn{1}{c}{\multirow{2}{*}{\shortstack{Property\\Response}}} & \multirow{2}{*}{Avg.} & \multicolumn{1}{c}{\multirow{2}{*}{\shortstack{Spatial\\Intelligence}}} & \multicolumn{1}{c}{\multirow{2}{*}{\shortstack{Strategic\\Reason}}} & \multicolumn{1}{c}{\multirow{2}{*}{\shortstack{Long-Horizon\\Plan}}} & \multicolumn{1}{c}{\multirow{2}{*}{\shortstack{Logic Puzzle\\Solving}}} & \multirow{2}{*}{Avg.} & \\
& & & & & & & & & & & \\
\midrule
\multicolumn{12}{c}{\cellcolor{pink!30}\textbf{Closed-source Models}} \\
\midrule
FLUX-Kontext-Pro & 47.35 & 47.16 & 44.37 & 41.44 & 45.00 & 49.12 & 48.58 & 51.16 & 40.49 & 46.52 & 45.77 \\
Seedream4.0 & 67.98 & 72.93 & 65.07 & 59.49 & 66.22 & 38.87 & 42.16 & 44.09 & 51.65 & 45.38 & 55.77 \\
Wan2.5 & 74.66 & 69.74 & 63.51 & 62.87 & 67.23 & 63.73 & 47.13 & 55.00 & 54.93 & 52.67 & 61.36 \\
Nano Banana & 77.10 & 78.88 & 71.86 & 74.70 & 75.22 & 66.74 & \textbf{56.11} & 56.83 & \textbf{64.91} & 60.39 & 68.26 \\
GPT-4o & \textbf{82.67} & \textbf{84.75} & \textbf{79.81} & \textbf{77.40} & \textbf{81.01} & \textbf{77.73} & 51.44 & \textbf{67.61} & 63.79 & \textbf{62.07} & \textbf{73.39} \\
\midrule
\multicolumn{12}{c}{\cellcolor{blue!10}\textbf{Open-source Models}} \\
\midrule
MagicBrush & 43.97 & 46.09 & 42.93 & 46.64 & 44.69 & 63.58 & 33.59 & 30.72 & 35.31 & 36.85 & 40.77 \\
Omnigen2 & 55.47 & 58.27 & 51.44 & 50.70 & 53.69 & \textbf{70.28} & 27.50 & 36.25 & 24.31 & 33.14 & 43.41 \\
Lumina-DiMOO & 52.84 & 52.31 & 50.31 & 50.83 & 51.44 & 61.23 & 36.96 & 39.57 & \textbf{53.09} & 45.61 & 48.54 \\
Step1X-Edit & 59.69 & 56.36 & 53.85 & 54.84 & 55.93 & 65.68 & 34.89 & 47.01 & 43.89 & 44.00 & 50.15 \\
Bagel-Think & 61.45 & 59.51 & 52.65 & 55.68 & 56.80 & 66.29 & 43.23 & 43.00 & 41.30 & 45.10 & 50.96 \\
DreamOmni2 & 63.52 & 58.35 & 52.68 & 54.02 & 56.64 & 72.42 & 42.17 & 48.80 & 48.09 & 48.98 & 52.81 \\
UniWorld-V2 & 72.96 & 69.37 & 63.65 & 61.69 & 66.55 & 49.27 & 40.00 & 53.41 & 37.53 & 43.19 & 54.87 \\
Qwen-Image-Edit & 75.68 & 73.03 & 70.59 & 64.67 & 70.95 & 56.73 & 36.63 & 48.80 & 37.68 & 41.92 & 56.52 \\
KnowThinker (Ours) & \textbf{80.74 }& \textbf{81.18} & \textbf{77.74} & \textbf{74.33} & \textbf{78.49} & 61.75 & \textbf{47.79} & \textbf{55.92} & 43.31 & \textbf{52.19} & \textbf{65.34} \\
\bottomrule
\end{tabular}%
\caption{In-domain quantitative comparisons on UniREditBench. GPT-4.1 is used as the evaluator. Best scores are in \textbf{bold}.}
\label{tab:unireditbench}
\end{table*}

\begin{table}[]
\centering
\vspace{-0.3cm}
\label{tab:conflict_resolution}
\setlength{\tabcolsep}{2.5pt}
\renewcommand{\arraystretch}{1.}
\resizebox{0.477\textwidth}{!}{%
\begin{tabular}{lccccc}
\toprule
\textbf{Conflict} & \textbf{Overall} & \textbf{Knowledge} & \textbf{Visual} & \textbf{Input} & \textbf{Rationale} \\
\textbf{Resolution} & \textbf{Score} & \textbf{Reasoning} & \textbf{Quality} & \textbf{Consistency} & \textbf{Alignment} \\
\midrule
Drop & 66.25 & 72.33 & 78.72 & 84.78 & 49.85 \\
Mean & 67.58 & 73.95 & 79.88 & 84.65 & 50.12 \\
PCGrad & 68.15 & 74.38 & 80.42 & 85.21 & 50.38 \\
Image Priority & 68.72 & 75.21 & 81.15 & 86.08 & 50.95 \\
Conservative & \textbf{69.73} & \textbf{76.58} & \textbf{82.31} & \textbf{86.59} & \textbf{52.20} \\
\bottomrule
\end{tabular}
}
\vspace{-0.15cm}
\caption{Impact of different conflict resolution of the CPAF.}
\vspace{-0.55cm}
\label{tab:conflict_resolution}
\end{table}

\vspace{0.2cm}
\section{Experiments and Analysis}

\subsection{Implementation Details}

\noindent
\textit{Training data, models, and setups.} For SFT, we fine-tune the Qwen3.5-27B~\cite{qwen35blog} or Qwen3-VL-8B~\cite{bai2025qwen3} reasoner on ExpertVerse-100K via LoRA with rank $r\!=\!16$ and scaling factor $\alpha\!=\!32$. Further, we construct the RFT dataset by sampling 10K high-quality instances via difficulty-aware distillation. We set $\tau_{\text{low}}\!=\!2$ and $\tau_{\text{high}}\!=\!4$, while $\tau_{\text{var}}$ is determined by thresholding probability distribution of rollout rewards. During RL, we freeze the vision encoder and first 32 layers, fine-tuning the remainder via LoRA. All RL experiments run on 40 NVIDIA A100 GPUs with Megatron-LM 5D. Qwen3.5-Plus evaluates the VLM-generated refined prompts and images from Qwen-Image-Edit-2511, yielding $r_\text{cot}$ and $r_\text{image}$. Full scoring prompts are provided \textcolor{blue}{\textbf{in Appendix}}. We set $\beta\!=\!0.3$ (Eq.~\ref{BRR}) in BRR, and $\alpha\!=\!0.5$ (Eq.~\ref{CPAF}) with a conservative conflict resolution strategy in CPAF.

\noindent
\textit{Evaluation Benchmarks.} Beyond ExpertVerse, we conduct holistic evaluations across multiple established benchmarks to assess diverse capabilities. These include RISEBench for temporal, causal, spatial, and logical reasoning; KRISBench for factual, conceptual, and procedural knowledge across seven dimensions; UniREditBench for game-world scenarios and multi-object interactions; and WiseEdit, which evaluates cognition- and creativity-informed editing via a three-stage pipeline of awareness, interpretation, and imagination.

\begin{figure*}[]
    \centering
    \caption{
    Qualitative comparisons of image editing task across the eight knowledge-intensive domains of ExpertVerse. Compared to state-of-the-art proprietary and open-source models, KnowThinker demonstrates remarkable superiority in both complex instruction following and high-fidelity visual quality.
    }
    \includegraphics[width=1\textwidth]{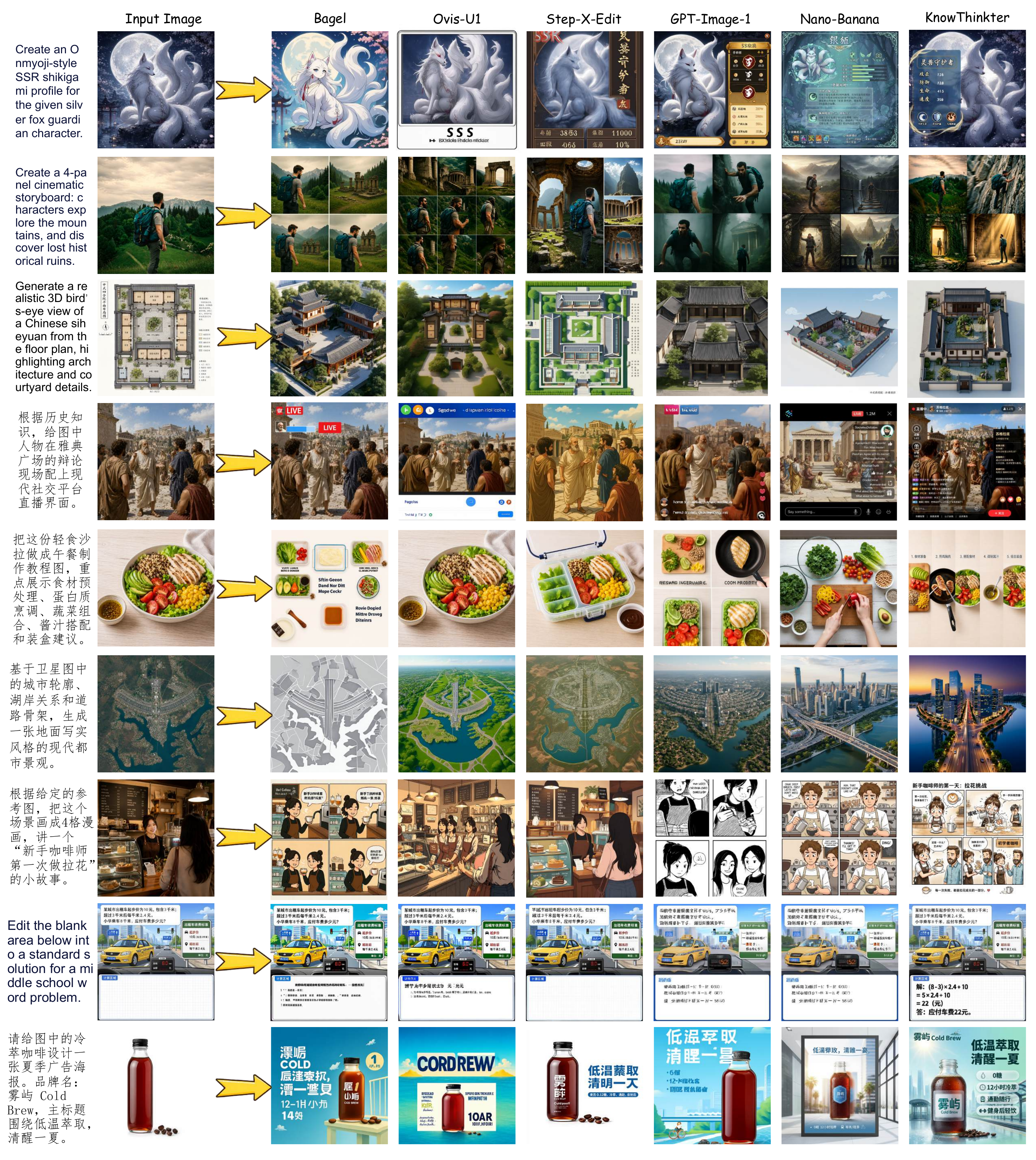}
    \label{fig:vis_comp}
\end{figure*}

\begin{figure*}[]
    \centering
    \caption{
    Qualitative comparisons of text-to-image generation across the knowledge-intensive ExpertVerse benchmark. Compared to state-of-the-art proprietary and open-source models, KnowThinker demonstrates remarkable superiority in both complex instruction following and high-fidelity visual quality.
    }
    \includegraphics[width=1\textwidth]{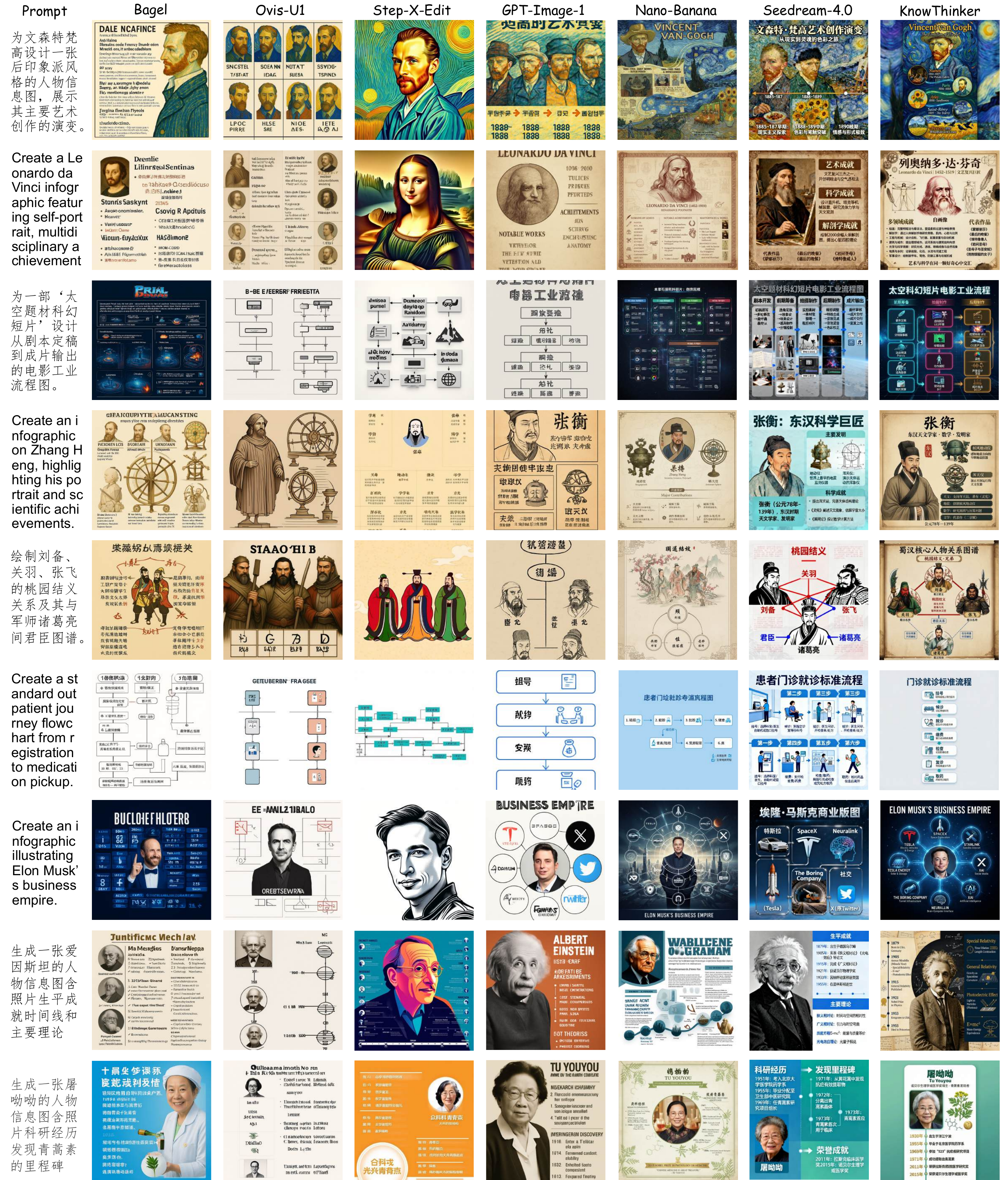}
    \label{fig:vis_comp_4}
\end{figure*}

\subsection{Comparison Results on Current Benchmarks}
\textbf{Performance on RISE-Bench.} In Tab.~\ref{tab:main_results}, KnowThinker-27B paired with frozen Qwen-Image-Edit-2511~\cite{wu2025qwen} editor achieves 40.0\% overall accuracy on RISEBench~\cite{zhao2026envisioning}, establishing a new state-of-the-art among community models by a substantial margin. It achieves a 20.6\% gain over base model (19.4\%) and remarkably outperforms joint-training baselines, \textit{e.g.}, Unified Thinker (28.9\%) and ThinkRL-Edit (29.7\%), strongly validating the efficacy of decoupling VLM reasoning from visual execution. By providing dense, cross-disciplinary reasoning traces, ExpertVerse-100K provides domain knowledge not merely for factual recall, guiding the model toward generalized reasoning, \textit{e.g.}, causal chaining and temporal sequencing. By addressing the cross-modal misalignment and gradient conflicts, BPPO prevents policy from prioritizing visual rewards over logical and causal reasoning chains. When evaluated against proprietary systems, KnowThinker exhibits competitive results, surpassing NanoBanana and approaching NanoBanana-Pro (47.2\%). While a gap remains in broader reasoning, we attribute this to frozen visual editor's limitations rather than planning deficits. 

\noindent
\textbf{Performance on KRIS-Bench.} Tab.~\ref{tab:kris_bench_results} demonstrates that KnowThinker achieves remarkable 83.02\% score on KRIS-Bench, outperforming all community models. It establishes a clear margin over recent reasoning-centric methods, \textit{e.g.}, ThinkRL-Edit (\textbf{\textcolor{green}{+11.37\%}}) and EditThinker (\textbf{\textcolor{green}{+11.11\%}}), validating effectiveness of knowledge-anchored reasoning. While prior methods struggle with procedural knowledge—which demands rigorous multi-step dynamic execution (\textit{e.g.}, ThinkRL-Edit scores merely 57.44\%)—KnowThinker achieves an exceptional 83.32\% (\textbf{\textcolor{green}{+25.88\%}}). This represents a remarkable leap from Qwen-Image-Edit-2511's procedural proficiency of 67.55\%, showing that our KnowThinker internalizes complex operational logic rather than relying on shallow factual recall. Further, KnowThinker delivers substantial gains across the conceptual and factual domains, surging by \textbf{\textcolor{green}{+21.50\%}} and \textbf{\textcolor{green}{+7.19\%}} over the base editor. Further, KnowThinker surpasses GPT-4o by 2.93\% and rivals the frontier NanoBanana-Pro (85.31\%), showing that decoupled reasoning can match 
end-to-end proprietary systems even with a frozen image editor.

\noindent
\textbf{Performance on UniREditBench.} As illustrated in Tab.~\ref{tab:unireditbench}, KnowThinker-27B achieves a remarkable overall score of 65.34\%, establishing itself as the strongest open-source model on the UniREditBench benchmark. Our method exhibits exceptional generalization capabilities across diverse environments: it achieves state-of-the-art results in \textit{Real World Scenario} tasks (78.49\% average) across all four sub-dimensions, while simultaneously delivering robust and leading performance in the highly challenging \textit{Game World Scenario} tasks (52.19\% average). This comprehensive superiority validates the broad applicability of our knowledge-anchored reasoning paradigm. Specifically, the cross-disciplinary knowledge traces from ExpertVerse-100K not only empower the model to handle physical commonsense in real-world edits but also endow it with the abstract reasoning capabilities required for virtual and rule-based environments. Notably, KnowThinker achieves the best open-source performance in \textit{Strategic Reason} (47.79\%) and \textit{Long-Horizon Plan} (55.92\%), demonstrating its exceptional proficiency in multi-step strategic planning and complex mathematical logic. Furthermore, our hierarchical cognitive planning mechanism significantly enhances the model's spatial and logical reasoning, yielding strong results in \textit{Spatial Intelligence} (61.75\%) and \textit{Logic Puzzle Solving} (43.31\%). The BPPO algorithm's cross-modal reward calibration plays a crucial role in these abstract tasks, ensuring that the model's cognitive plans for intricate spatial layouts and abstract rule-based puzzles are faithfully and accurately translated into visual execution. It underscores that KnowThinker is a versatile cognitive agent capable of seamlessly mastering both physical commonsense and abstract logical reasoning.

\noindent
\textbf{Performance on WiseEdit-Bench.} 
As illustrated in Tab.~\ref{tab:wiseedit_main}, KnowThinker achieves an overall score of 74.4 on the standard WiseEdit benchmark, outperforming the leading open-source model (DreamOmni2) by 13.8 points and approaching the proprietary GPT-image-1 (77.6). Notably, on the \textit{Imagination Task}, it scores 77.7, surpassing all baselines including the closed-source Nano Banana Pro (76.6). This indicates that our decoupled cognitive planning effectively mitigates semantic drift during complex synthesis. Furthermore, on the \textit{Awareness} and \textit{Interpretation} tasks, KnowThinker yields significant gains in Instruction Following ($\mathcal{IF}$) and Knowledge Fidelity ($\mathcal{KF}$) over its base editor (Qwen-Image-Edit), demonstrating that ExpertVerse's reasoning traces successfully ground the generation in factual constraints. The advantage is further amplified on the challenging WiseEdit-Complex benchmark (Tab.~\ref{tab:WiseEdit-Complex}), which requires multi-image grounding and cross-lingual reasoning. KnowThinker achieves an overall score of 82.4, surpassing all open-source competitors (capped at 54.8) and outperforming proprietary systems like Nano Banana Pro (77.1) and Seedream 4.0 (71.4). This performance is consistent across English (81.8) and Chinese (82.9) subsets, delivering high $\mathcal{IF}$ (89.3 / 87.2) and $\mathcal{KF}$ (89.7) alongside strong Visual Quality (94.2 in Chinese). These results confirm that BPPO resolves gradient conflicts in multi-constraint editing, ensuring instruction adherence without compromising visual coherence, and validating KnowThinker as a robust, knowledge-driven agent for complex multi-modal tasks.

\begin{table}[t]
\centering
\vspace{-0.3cm}
\setlength{\tabcolsep}{1.8pt}
\small
\begin{tabular}{ccccccc}
\toprule
\textbf{ID}  & \textbf{SFT} & \textbf{BRR} & \textbf{CPAF} & \textbf{RISE-Bench} & \textbf{KRIS-Bench} & \textbf{ExpertVerse} \\ \midrule
(a) &  \textcolor{red}{\ding{55}}   &  \textcolor{red}{\ding{55}}   &  \textcolor{red}{\ding{55}}    & 33.34      & 75.42      & 54.17       \\
(b) & \textcolor{green}{\ding{51}}   &   \textcolor{red}{\ding{55}}  &  \textcolor{red}{\ding{55}}    & 37.81      & 80.09      & 66.22       \\
(c) &  \textcolor{red}{\ding{55}}   & \textcolor{green}{\ding{51}}   &   \textcolor{red}{\ding{55}}   & 34.12      & 76.15      & 65.28       \\
(d) &  \textcolor{red}{\ding{55}}   &  \textcolor{red}{\ding{55}}   & \textcolor{green}{\ding{51}}    & 33.95      & 75.88      & 64.94       \\
(e) &   \textcolor{red}{\ding{55}}  & \textcolor{green}{\ding{51}}   & \textcolor{green}{\ding{51}}    & 35.27      & 77.63      & 66.81       \\
(f) & \textcolor{green}{\ding{51}}   & \textcolor{green}{\ding{51}}   &   \textcolor{red}{\ding{55}}   & 39.24      & 82.41      & 68.85       \\
(g) & \textcolor{green}{\ding{51}}   &  \textcolor{red}{\ding{55}}   & \textcolor{green}{\ding{51}}    & 38.67      & 81.83      & 67.92       \\
(h) & \textcolor{green}{\ding{51}}   & \textcolor{green}{\ding{51}}   & \textcolor{green}{\ding{51}}    & \textbf{40.00}      & \textbf{83.02}      & \textbf{69.73}       \\ \bottomrule
\end{tabular}
\vspace{-0.17cm}
\caption{Ablation study on core components of KnowThinker.}
\vspace{-0.7cm}
\label{tab:main_ablation}
\end{table}

\vspace{-0.25cm}
\subsection{Benchmarking Results on ExpertVerse} 
\vspace{-0.05cm}
Tab.~\ref{tab:domain_performance} evaluates models across eight cross-disciplinary domains on ExpertVerse, revealing their strengths, weaknesses, and recent advancements. Beyond basic visual comprehension, executing complex editing tasks requires rigorous cross-disciplinary knowledge reasoning to bridge the gap between abstract semantic constraints and high-fidelity visual synthesis. However, the performance gap between open-source and closed-source models remains significant. While proprietary models (\textit{e.g.}, GPT-Image-2 and Nano Banana Pro) dominate T2I tasks, they exhibit notable performance decay in I2I editing and fine-grained quantitative reasoning (Data\&Info. and Sci.\&Edu.). Open-source baselines suffer from severe knowledge blind spots, performing sub-optimally in multi-hop reasoning domains (\textit{e.g.}, Geo.\&His). Fig.~\ref{fig:vis_comp} and Fig.~\ref{fig:vis_comp_4} highlights that knowledge-intensive reasoning remains a critical bottleneck, as current models (\textit{e.g.}, Ovis-U1, Bagel) struggle to leverage internal world knowledge to execute editing instructions. The KnowThinker bridges the open-closed source divide, surpassing GPT-Image-2 in specific I2I scenarios (\textit{e.g.}, 92.38 in Fash. \& Beau.\!vs.\!75.11). These gains arise from the synergy between the multi-capability training dataset for cross-disciplinary knowledge internalization and knowledge-intensive reasoning traces for editing planning.

\begin{figure}[]
    \centering
    \caption{
    Impact of refined instructions across the evaluation metrics, showing zero-shot transferability of KnowThinker.
    }
    \vspace{-0.31cm}
    \includegraphics[width=0.474\textwidth,height=0.304\textwidth]{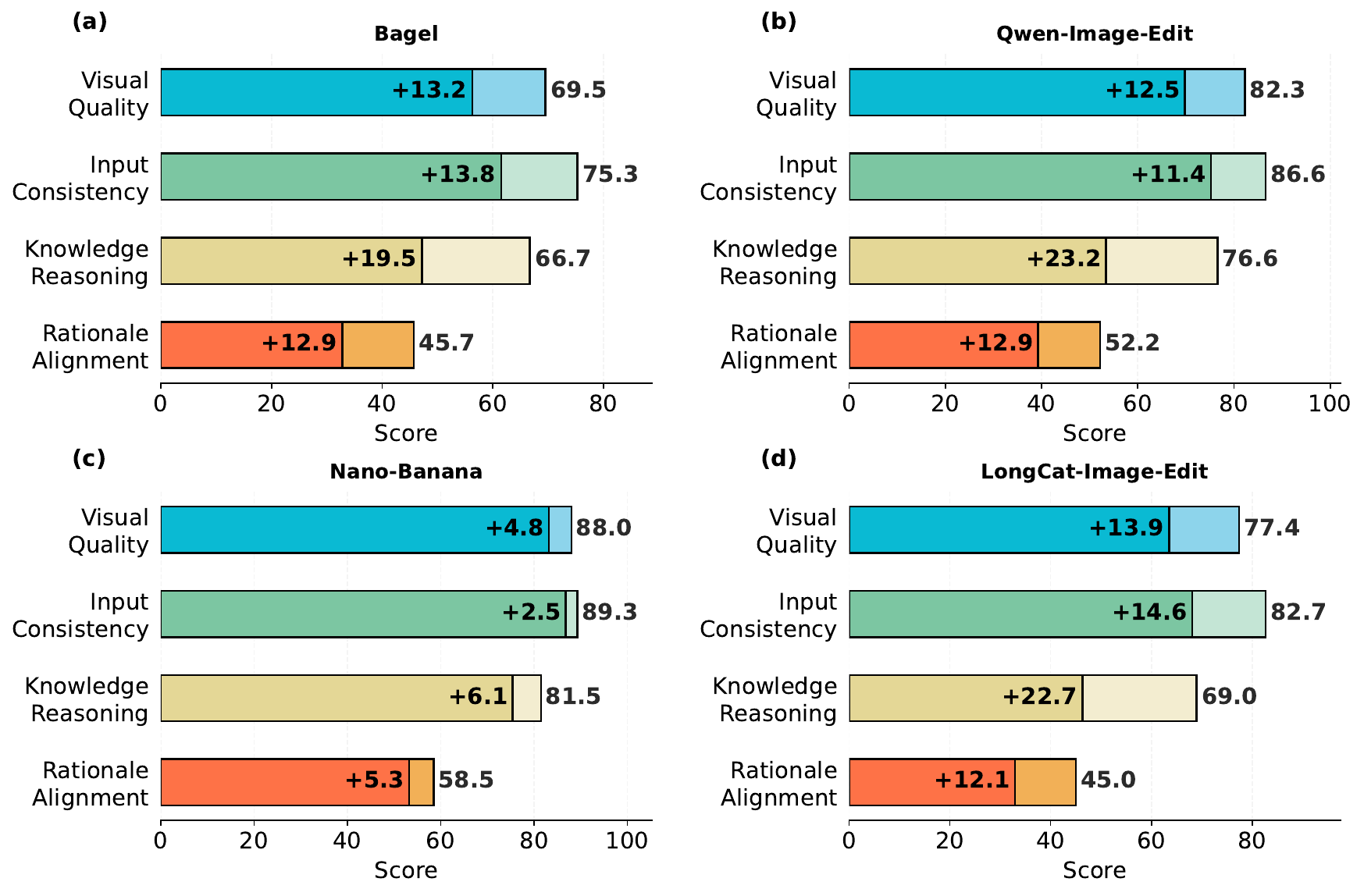}
    \label{fig:compard_with_others}
    \vspace{-0.91cm}
\end{figure}

\subsection{In-Depth Ablation Studies and Analysis}
\textbf{Effectiveness of training strategy.} We provide several variants to validate effectiveness of our training strategy in Tab.~\ref{tab:main_ablation}. Comparing (a) to (e) shows that SFT cold-start yields substantial gains (\textbf{\textcolor{green}{+12.05}} on ExpertVerse, \textbf{\textcolor{green}{+4.67}} on KRIS-Bench), whereas applying BPPO without SFT results in sub-optimal performance. To explain, SFT is indispensable for internalizing cross-disciplinary knowledge priors and establishing the structured cognitive planning required for RL exploration. Incorporating BRR (f) or CPAF (g) individually yields substantial performance surges, such as \textbf{\textcolor{green}{+2.63}} (to 68.85) and \textbf{\textcolor{green}{+1.70}} (to 67.92) on ExpertVerse, validating their efficacy in aligning visual execution with expert-level reasoning in a unified multi-objective optimization framework. Ultimately, the full training paradigm (h) synergizes these mechanisms, establishing new state-of-the-art results across all benchmarks.

\noindent
\textbf{Impact of instruction rewriting.} Fig.~\ref{fig:compard_with_others} validates the zero-shot transferability of KnowThinker across four editing backbones. The instruction refinement of KnowThinker delivers universal performance gains, most prominently in \textit{Knowledge Reasoning} and \textit{Rationale Alignment} (\textit{e.g.}, \textbf{\textcolor{green}{+23.2}} and \textbf{\textcolor{green}{+12.9}} on Qwen-Image-Edit). We conclude that explicit cognitive planning and knowledge-anchored thinking rectify logical deficits and factual hallucinations inherent in editors. The structured reasoning directly cascades into visual enhancements, delivering substantial absolute gains in \textit{Visual Quality} (up to \textbf{\textcolor{green}{+13.9}}) and \textit{Input Consistency} (up to \textbf{\textcolor{green}{+14.6}}). It confirms that explicit intermediate reasoning steps underpins high-fidelity editing, positioning KnowThinker as a general reasoning edit engine.

\noindent
\textbf{Impact of different conflict resolution.} Tab.~\ref{tab:conflict_resolution} evaluates the impact of various conflict resolutions in CPAF. Naive baselines like \textit{Drop} (halting updates when conflicts) and \textit{Mean} (direct gradient averaging) suffer from optimization stagnation, yielding sub-optimal scores of 66.25 and 67.58. While \textit{PCGrad}~\cite{yu2020gradient} mitigates conflicts by projecting gradients to eliminate conflicting components (68.15), and \textit{Image Priority} exclusively follows image rewards to maximize Visual Quality (81.15), the latter severely compromises Rationale Alignment (50.95) and Knowledge Reasoning (75.21). In contrast, \textit{Conservative} strategy selects gradient with smaller absolute magnitude, thus preventing extreme parameter updates. This mechanism achieves a pareto-optimal trade-off, yielding the highest overall score and validating its efficacy in aligning image editing with knowledge-intensive reasoning.

\noindent\textbf{Effectiveness of KnowThinker as a knowledge-intensive prompt rewriter.}
As shown in Tab.~\ref{tab:thinker_detailed_default_editor}, we ablate the VLM thinker by freezing visual editor, making the refined prompt the sole variable. Directly feeding raw instructions yields an average score of 38.81, indicating that knowledge-intensive requests are too abstract for the editor to execute reliably. In contrast, KnowThinker (27B) achieves an average score of 69.73, improving Knowledge Reasoning, Visual Quality, Input Consistency, and Rationale Alignment by 32.20, 22.16, 18.25, and 19.16 points, respectively. Notably, KnowThinker achieves a state-of-the-art \textit{Input Consistency} of 86.59, surpassing all open-source baselines and outperforming proprietary models like Claude-4.8-Opus (76.79), GPT-5.5 (76.47), and Gemini-3.1-Pro (67.48). This result suggests that our framework effectively mitigates the ``\textit{over-synthesis}'' hallucination, preserving the reference image's structural integrity. Compared to the open-source model, KnowThinker outperforms Qwen3.5-27B (\textcolor{green}{\textbf{+2.85}}) and the MoE-based Qwen3.5-35B-A3B (\textcolor{green}{\textbf{+6.58}}). Although the Qwen3.5-35B-A3B model shows a slight edge in Input Consistency (82.66), it lags in Knowledge Reasoning (67.80) and Rationale Alignment (48.77), whereas KnowThinker achieves a balanced performance across all dimensions. While a marginal gap remains in Knowledge Reasoning against GPT-5.5 (81.35) and Claude-4.8-Opus (79.91), the overall results validate that the knowledge-intensive supervision from ExpertVerse-100K and the conflict-aware optimization of BPPO enable KnowThinker to act as an effective prompt rewriter, bridging the gap between implicit requests and frozen image editors.

\begin{table}[]
\centering
\vspace{-0.15cm}
\setlength{\tabcolsep}{3.2pt}
\renewcommand{\arraystretch}{1.05}
\resizebox{\linewidth}{!}{%
\begin{tabular}{lccccc}
\toprule
\textbf{Thinker} &
\textbf{Avg} &
\textbf{Knowledge} &
\textbf{Visual} &
\textbf{Input} &
\textbf{Rationale} \\
&
\textbf{Score} &
\textbf{Reasoning} &
\textbf{Quality} &
\textbf{Consist.} &
\textbf{Alignment} \\
\midrule
Raw Instruction
& 38.81 & 44.38 & 60.15 & 68.34 & 33.04 \\
\midrule
Claude-4.8-opus
& 75.14 & 79.91 & 86.23 & 76.79 & 62.13 \\

Gemini-3.1-pro
& 73.92 & 80.72 & 89.19 & 67.48 & 58.14 \\

GPT-5.5
& 76.97 & 81.35 & 87.34 & 76.47 & 65.76 \\
\midrule
Qwen3.5-27B
& 66.88 & 72.64 & 81.37 & 70.58 & 51.57 \\
Qwen3.5-35B-A3B
& 64.50 & 69.31 & 77.75 & 83.05 & 50.34 \\

KnowThinker (27B)
& \textbf{69.73} & \textbf{76.58}
& \textbf{82.31} & \textbf{86.59}
& \textbf{52.20} \\
\bottomrule
\end{tabular}%
}
\vspace{-0.1cm}
\caption{Comparison of VLM thinkers using Qwen-Image-Edit as visual editor. All thinkers receive identical images and user instructions, and only refined prompts are varied.}
\vspace{-0.4cm}
\label{tab:thinker_detailed_default_editor}
\end{table}

\section{Conclusion}
We delve into knowledge-intensive image synthesis, which demands expert-level cross-disciplinary, structured cognitive plans, and domain-anchored knowledge rationales. We introduce a knowledge-intensive benchmark ExpertVerse alongside the large-scale ExpertVerse-100K corpus for reasoning-based image generation. We propose KnowThinker, a thinker-centric VLM reasoner, decoupling cognitive planning from a frozen visual editor to decompose instructions into knowledge-anchored visual plans. To train KnowThinker, we develop a novel RL algorithm BPPO, which combines BRR for cross-modal reward calibration and CPAF for the advantage-level objective fusion. KnowThinker achieves state-of-the-art records on all reasoning editing benchmarks. ExpertVerse illuminates reasoning bottlenecks of current models, offering a paradigm to catalyze the evolution of knowledge-aware visual synthesis.


\bibliography{aaai2027}


\end{document}